\documentclass[journal]{IEEEtran}
\ifCLASSINFOpdf
\else
\fi
\usepackage{graphicx}
\usepackage{cite}
\usepackage{amssymb}
\usepackage{amsmath}
\usepackage{color}
\usepackage{enumerate}
\usepackage{epsf}
\usepackage{subfigure}
\usepackage{mathrsfs}

\newtheorem{assumption}{Assumption}

\newcommand{\Tr}{\mathrm{Tr}}

\newcommand{\vect}{\mathrm{vec}}
\newcommand{\veq}{\mathrel{\phantom{=}} }

\long\def\comment#1{}

\begin{document}
%
\title{Random Euler Complex-Valued Nonlinear Filters
}
%
%

\graphicspath{{figures/}}

\author{Jiashu Zhang, Sheng Zhang, and Defang Li
\thanks{Manuscript received **, 2017. This work was supported by the National Science Foundation of China (Grants 61671392, U1562218).}
\thanks{The authors are with the School of Information Science and Technology,
Southwest Jiaotong University, Chengdu 611756, China (e-mail: jszhang@swjtu.edu.cn, dr.s.zhang@ieee.org, Ldf125@swjtu.edu.cn).
%
}
}

\maketitle

\begin{abstract}
Over the last decade, both the neural network and kernel adaptive filter have successfully been used for nonlinear signal processing. However, they suffer from high computational cost caused by their complex/growing network structures. In this paper, we propose two random Euler filters for complex-valued nonlinear filtering problem, i.e., linear random Euler complex-valued filter (LRECF) and its widely-linear version (WLRECF), which possess a simple and fixed network structure. The transient and steady-state performances are studied in a non-stationary environment. The analytical minimum mean square error (MSE) and optimum step-size are derived. Finally, numerical simulations on complex-valued nonlinear system identification and nonlinear channel equalization are presented to show the effectiveness of the proposed methods.
\end{abstract}


%
\IEEEpeerreviewmaketitle

\section{Introduction}
%
%
%
%
With the development of adaptive filtering, complex-valued adaptive filter has found applications in diverse fields of radar imaging, fourier analysis, mobile communications, seismics, estimation of direction of arrival and beamforming \cite{IEEEhowto1,IEEEhowto2,IEEETS2010}. In modeling and identification of complex-valued nonlinear systems, traditional linear adaptive filtering techniques suffer from poor performance. Examples for such situations include nonlinear system identification, nonlinear channel equalization. In order to model nonlinear systems, serval methods have been proposed in the last half century \cite{IEEETS2010,IEEEDV2009,IEEEWJS2010,IEEEHJ2009,IEEEMD2013,IEEEAG2014,IEEESW2017}, which include the neural networks, polynomial, spline and Fourier filters, to just mention a few. 

In order to directly process complex values by neural networks, the complex-valued neural network (CVNN) have been developed \cite{IEEEDJ2001,IEEEAH2006,IEEEHS1991, IEEEHT2008}, where the splitting-complex and fully-complex activation functions are used. The major drawback of the CVNNs is the heavy computational complexity. Several different types of CVNNs have been presented in \cite{IEEEDJ2001}, such as multiplayer percetron (MLP) networks, radial basis function (RBF) networks, and recurrent neural networks (RNN). In \cite{IEEEYBMJ2011}, the echo state network for complex noncircular signals was proposed, which separates the RNN architecture into two constituent components: a recurrent architecture and a memoryless output layer. With a complex-chebyshev expansion for the input signal, the complex-chebyshev functional-link network (CCFLN) was designed \cite{IEEEMJYW2012}, which is a linear filtering of the expanded signal in the higher dimensional space.

Based on the reproducing kernel Hilbert space (RKHS) theory, the kernel adaptive filters (KAFs) were developed in \cite{IEEEWJS2010}, which maps the original input space into an infinite dimensional RKHS with a specific kernel. When the kernel is chosen as Gaussian kernel, the KAF is the growing RBF network. Over the real kernel filter, serval adaptive algorithms were proposed in \cite{IEEEKSI2009,IEEEWJCJ2012,IEEEWJCJ2004,IEEEYSR2004}, such as the kernel least mean square (KLMS), kernel affine projection, kernel recursive least-squares, kernel projected subgradient methods. Using the \textit{complexification} of real RKHSs, or complex reproducing kernels, the complex kernel adaptive filtering has been introduced in \cite{IEEEPKS2012}. With the wide-linear model, further enhancements to the complex-valued/quaternion-valued kernel approach can be found in \cite{IEEEFAD2012,IEEEPSM2012,IEEETT2015}. However, the order of these kernel filters grows linearly with the number of input data. To overcome this severe drawback, several low-complexity techniques have been developed in \cite{IEEEWIJP2009,IEEEBSSJ2012,IEEEBSPJ2012,IEEEANP2012,IEEEWJCJ2014}, such as the sparse KLMS, quantized KLMS, KLMS with $l_1$-norm regularization.

Recently, according to Bochner's theorem, Rahimi and Recht suggested a popular approach, i.e., random fourier features, to approximate the real kernel evaluation in KAFs \cite{IEEEAB2007}. Based on the random fourier features, the random fourier filtering (RFF) has been proposed in \cite{IEEEPSS2016}, where the original input data is mapped to a finite dimensional space. Thus, compared with the KAF, it enables learning of nonlinear functions in an efficient fashion. In \cite{IEEEPSS2016}, the mean square (LMS) and recursive least squares (RLS) were developed into the RFF. Furthermore, a distributed RFF was presented for networks in \cite{IEEEPSS2017}. Unfortunately, these RFFs only deal with real-valued nonlinear systems.


%
In this paper, we propose two random Euler complex-valued filters to deal with the complex-valued nonlinear filtering problem.
Firstly, based on the \textit{complexification} of real RKHSs and Bochner's theorem, a detailed derivation of the linear random Euler complex-valued filter (LRECF) is presented. Then,
employing the widely-linear model and the Wirtinger's derivative, the widely-linear random Euler complex-valued filter (WLRECF) is designed. Due to the fixed network structure, the proposed two schemes enjoy low computational complexity compared to the kernel filter. Theoretical analysis on the mean stability and mean-square convergence of the proposed methods is performed in a non-stationary environment modeled by a random-walk model. From these results, the closed-form expression for the steady-state mean square error (MSE) is obtained, which indicates that there is an optimum step-size in the non-stationary environment. Finally, experiments are conducted to evaluate the performance of the proposed filters, including complex-valued nonlinear system identification and nonlinear channel equalization.

The rest of this paper is organized as follows. In Section~\ref{sec:Rv}, a brief review of the RFF is presented. Section~\ref{sec:Pro} provides the derivation of the LRECF and WLRECF. The mean and mean-square behaviors are analyzed in Section~\ref{sec:Per}. Section~\ref{sec:MON} presents Monte Carlo simulations. Finally, conclusions are drawn in Section~\ref{sec:Co}. In this paper, matrices are represented by boldface capital letters, and all vectors are column vectors denoted by boldface lowercase letters. The other symbols are listed as follows:

\begin{enumerate}[xxxxxxxx]
\item[$(\cdot)^T$~~]  transpose operator;

\item[$(\cdot)^*$~~]  conjugate operator;

\item[$(\cdot)^H$~~]  Hermitian transpose operator;

\item[$\lambda_{\max}(\cdot)$~~]  largest eigenvalue of a matrix;

\item[$\Tr(\cdot)$~~]  trace of a matrix;

\item[$\mathbf{I}$~~]  identity matrix with appropriate dimension;

\item[$\mathbf{0}$~~]  zero vector with appropriate dimension;

\item[$\mathbf{A}\otimes \mathbf{B}$~~]  Kronecker product of two matrices $\mathbf{A}$ and $\mathbf{B}$;
%

\item[$\vect(\cdot)$~~]  column vector formed by stacking the columns of a matrix;

\item[$|\cdot|$~~]  absolute value of a complex number;

\item[$\textrm{real}(\cdot)$~~] real part of complex number;

\item[$\textrm{imag}(\cdot)$~~] imaginary part of complex number;

\item[$E\{\cdot\}$~~]  expectation operator.
\end{enumerate}
\begin{figure}[!tp]
\centering
\hspace*{-1em}
\includegraphics[height=45mm]{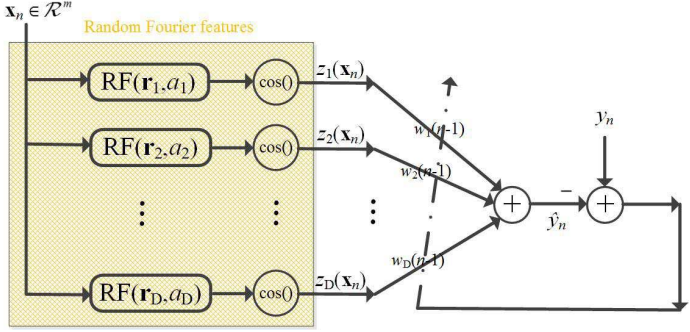}
\caption{The real random Fourier filter.}
\label{RRFNF1}
\end{figure}
\section{Review of Real Random Fourier Filter\label{sec:Rv}}
Consider a continuous nonlinear input-output mapping,
\begin{align}
y=f(\textbf{x})
\end{align}
where $\textbf{x}\in \mathcal{F}^m$ is the \textit{m}-dimensional vector\footnote{$\mathcal{F}$ is a general field, which can be either $\mathcal{R}$ or $\mathcal{C}$.}, and $y\in \mathcal{F}$ is the output signal. Based on the training sequences of the form $\{(\textbf{x}_n,y_n),n=1,2,\cdots\}$, the goal of the learning tasks is to learn the non-linear input-output dependence.

In the case of real Hilbert spaces (i.e., $\mathcal{F}=\mathcal{R}$), the real random Fourier nonlinear filter algorithm is
\begin{align}
\textbf{w}_n= \textbf{w}_{n-1}+\mu\textbf{z}(\textbf{x}_n)e_n
\end{align}
where $\mu$ is the step-size and $e_n=y_n-\textbf{z}^T(\textbf{x}_n)\textbf{w}_{n-1}$ with $\textbf{w}_{n-1}$ being a weight vector for the random Fourier
features vector $\textbf{z}(\textbf{x}_n)$. In \cite{IEEEAB2007}, it gives two such embeddings about the random Fourier
features $\textbf{z}(\textbf{x})$:
\begin{align}\label{ekyuer5}
\textbf{z}(\textbf{x})=\sqrt{\frac{1}{D}}\left[
	\begin{array}{c}
	 \sin(\textbf{r}_1^T\textbf{x})       \\ 
	 \cos(\textbf{r}_1^T\textbf{x})         \\
     \vdots  \\
     \sin(\textbf{r}_{2D}^T\textbf{x})     \\
     \cos(\textbf{r}_{2D}^T\textbf{x})
	\end{array}
\right],
\sqrt{\frac{2}{D}}\left[
	\begin{array}{c}
	 \cos(\textbf{r}_1^T\textbf{x}+a_1)       \\ 
	 \cos(\textbf{r}_2^T\textbf{x}+a_2)         \\
     \vdots  \\
     \cos(\textbf{r}_D^T\textbf{x}+a_D)
	\end{array}
\right],
\end{align}
where $\textbf{r}_i$ is drawn from a Gaussian distribution with zero mean and covariance matrix $\sigma^2\mathbf{I}$, $a_i$ is the uniform distribution
on $[0, 2\pi]$. Fig.~\ref{RRFNF1} shows the RFF with later embedding in (\ref{ekyuer5}).  As can be seen, in the RFF, the original data $\textbf{x}_n\in \mathcal{R}^m$ is transformed to a high dimensional feature space, via a map, $\textbf{z}(\textbf{x}_n)\in \mathcal{R}^D$.
\section{Proposed Random Euler Complex-Valued Nonlinear Filter\label{sec:Pro}}

In this section, we are interested on complex Hilbert spaces, i.e., $\mathcal{F}=\mathcal{C}$, and will design the LRECF and WLRECF.
Let $\textbf{z}\in\mathcal{C}^m$ and $\textbf{z}=\textbf{z}_1+i\textbf{z}_2$, $\textbf{z}_1,\textbf{z}_2\in\mathcal{R}^m$. We adopt the \textit{complexification} methodology of real RKHSs \cite{IEEEPKS2012},
\begin{align}
\Phi_c(\textbf{z})=&\Phi(\textbf{z}_1)+i\Phi(\textbf{z}_2)\notag\\
=&\kappa_\mathcal{R}(\cdot, [\textbf{z}_1,\textbf{z}_2])+i\kappa_\mathcal{R}(\cdot, [\textbf{z}_1,\textbf{z}_2])
\end{align}
where $\kappa_\mathcal{R}(\cdot, \cdot)$ is chosen as a real Gaussian kernel.

\subsection{Linear Random Euler Complex-Valued Filter}
By the use of the cost function, $\frac{1}{2}|y_n-\textbf{w}^H\Phi_c(\textbf{x}_n)|^2$, and the gradient descent method, at time $n$, the weight-update equation gives
$\textbf{w}_n=\sum_{i=1}^{n}\mu e_i^*\Phi_{c}(\textbf{x}_i)$\footnote{This is well-known complex-valued KLMS (CKLMS) via complexification of real kernels in \cite{IEEEPKS2012}.}, where the initial estimate is assumed to be zero and $e_n=y_n-\textbf{w}^H_{n-1}\Phi_{c}(\textbf{x}_n)$. The system's output $\hat{y}_n$, at time $n$, can be estimated as
\begin{align}
\label{1235f24a}
\hat{y}_n&=\textbf{w}^H_{n-1}\Phi_c(\textbf{x}_n)\\
&=2\sum_{i=1}^{n-1}\alpha_i\kappa_\mathcal{R}\left([\textrm{real}(\textbf{x}_n),~\textrm{imag}(\textbf{x}_n)],[\textrm{real}(\textbf{x}_i),~\textrm{imag}(\textbf{x}_i)]\right)\notag
\end{align}
where $\alpha_i=\mu e_i$. According to Bochner's theorem, we can have
\begin{align}
\label{1235fa}
&\kappa_\mathcal{R}\left([\textrm{real}(\textbf{x}_n),~\textrm{imag}(\textbf{x}_n)], [\textrm{real}(\textbf{x}_i),~\textrm{imag}(\textbf{x}_i)]\right)\notag\\
=&E_\textbf{c}\{\zeta_\textbf{c}([\textrm{real}(\textbf{x}_n),~\textrm{imag}(\textbf{x}_n)])\zeta_\textbf{c}^*([\textrm{real}(\textbf{x}_i),~ \textrm{imag}(\textbf{x}_i)])
\}
\end{align}
where $\zeta_\textbf{c}([\textrm{real}(\textbf{x}_n), \textrm{imag}(\textbf{x}_n)])=e^{j\textbf{c}^T[\textrm{real}(\textbf{x}_n); \textrm{imag}(\textbf{x}_n)]}$, and the random vector \textbf{c} is drawn from a probability distribution\footnote{It is actually the multivariate Gaussian distribution with zero mean and covariance matrix $\sigma^2\mathbf{I}$ \cite{IEEEPSS2017}.} that is the Fourier transform of the Gaussian kernel.

We choose a sample average to approximate (\ref{1235fa}) using \textit{D} random vectors $\{\textbf{c}_1,\textbf{c}_2,\cdots,\textbf{c}_D\}$,
\begin{align}
\label{1235fahrt}
&E_\textbf{c}\{\zeta_\textbf{c}([\textrm{real}(\textbf{x}_n), \textrm{imag}(\textbf{x}_n)])\zeta_\textbf{c}^*([\textrm{real}(\textbf{x}_i), \textrm{imag}(\textbf{x}_i)])
\}\notag\\
\approx &\frac{1}{D}\sum_i^D\zeta_{\textbf{c}_i}([\textrm{real}(\textbf{x}_n), \textrm{imag}(\textbf{x}_n)])\zeta_{\textbf{c}_i}^*([\textrm{real}(\textbf{x}_i), \textrm{imag}(\textbf{x}_i)])
\end{align}

Substituting (\ref{1235fa}) and (\ref{1235fahrt}) into (\ref{1235f24a}), the estimate of the filtering output can be approximated as
\begin{align}
\label{1235f24ahst}
\hat{y}_n=\textbf{u}^H_{n-1}\textbf{z}_c(\textbf{x}_n)
\end{align}
where $\textbf{u}_{n}=\sum_{i=1}^{n}\alpha_i^*\textbf{z}_c(\textbf{x}_i)$ with $\textbf{z}_c(\textbf{x}_i)$ being
\begin{align}\label{ekyu5}
\textbf{z}_c(\textbf{x}_i)=
\sqrt{\frac{2}{D}}\left[
	\begin{array}{c}
	 e^{j\textbf{c}^T_1[\textrm{real}(\textbf{x}_i); \textrm{imag}(\textbf{x}_i)]}       \\ 
	  e^{j\textbf{c}^T_2[\textrm{real}(\textbf{x}_i); \textrm{imag}(\textbf{x}_i)]}         \\
     \vdots  \\
      e^{j\textbf{c}^T_D[\textrm{real}(\textbf{x}_i); \textrm{imag}(\textbf{x}_i)]}
	\end{array}
\right]
\end{align}

In (\ref{1235f24ahst}), $\textbf{u}_{n-1}$ can be seen as a weight vector for the random features vector $\textbf{z}_c(\textbf{x}_n)$, which can be rewritten as
\begin{align}
\label{1235f24ahsoput}
\textbf{u}_{n}=\textbf{u}_{n-1}+\mu e_n^*\textbf{z}_c(\textbf{x}_n)
\end{align}
where the initial weight vector is assumed to be zero.

\begin{figure}[!t]
\centering
\hspace*{-1em}
\includegraphics[height=45mm]{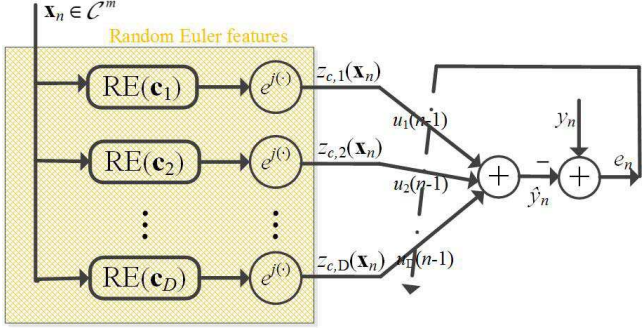}
\caption{The linear random Euler complex-valued filter.}
\label{LCRENF}
\end{figure}

Because of the Euler representation in (\ref{ekyu5}), we coin this approach as LRECF. Fig.~\ref{LCRENF} illustrates its architecture. As can be seen, the LRECF has fixed network structure, which is obviously different from the growing structure of the CKLMS.

\subsection{Widely-Linear Random Euler Complex-Valued Filter}
Inspired by (\ref{1235f24ahst}) and using the widely-linear model\footnote{The widely-linear model enables the processing of the
noncircular complex-valued signals, which provide improved performance than the conventional linear model \cite{IEEEBP1995,IEEEYCD2010,IEEEACNR2010,IEEEYLCH2015}.}, the estimator form of the nonlinear filter is written as
\begin{align}
\label{1235f24oput}
\hat{y}_n=&\textbf{u}^H\textbf{z}_c(\textbf{x}_n)+\textbf{v}^H\textbf{z}_c^*(\textbf{x}_n)
\end{align}

To design a filter $\{\textbf{u},~\textbf{v}\}$, we establish the following cost function
\begin{align}
\label{12put}
\mathcal{L}(e_n)=\frac{1}{2}|y_n-\textbf{u}^H\textbf{z}_c(\textbf{x}_n)-\textbf{v}^H\textbf{z}_c^*(\textbf{x}_n)|^2
\end{align}
where the error signal is $e_n=y_n-\hat{y}_n$.

\begin{figure}[!t]
\centering
\hspace*{-1em}
\includegraphics[height=77mm]{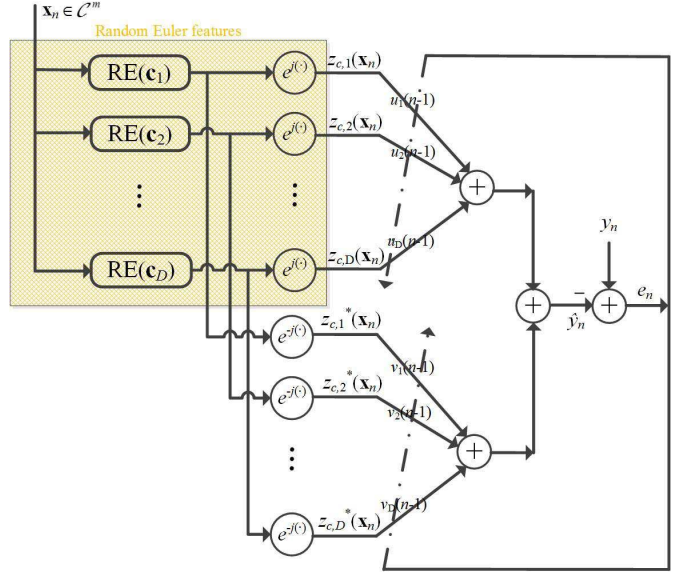}
\caption{The widely-linear random Euler complex-valued filter.}
\label{WLCRENF}
\end{figure}

Using the stochastic gradient adaptation and the Wirtinger's derivative with respect to $\{\textbf{u},\textbf{v}\}$, as follows:
\begin{align}
\nabla_{\textbf{u}} \mathcal{L}(e_n)=2\frac{\partial \mathcal{L}(e_n)}{\partial \textbf{u}^*}
=\frac{\partial \mathcal{L}(e_n)}{\partial \textbf{u}_r}+i\frac{\partial \mathcal{L}(e_n)}{\partial \textbf{u}_i}
\end{align}
and
\begin{align}
\nabla_{\textbf{v}} \mathcal{L}(e_n)=2\frac{\partial \mathcal{L}(e_n)}{\partial \textbf{v}^*}
=\frac{\partial \mathcal{L}(e_n)}{\partial \textbf{v}_r}+i\frac{\partial \mathcal{L}(e_n)}{\partial \textbf{v}_i}
\end{align}
we get the update equation for the weight vector
\begin{align}
\label{12put1}
\textbf{u}_n=&\textbf{u}_{n-1}-\mu \nabla_{\textbf{u}} \mathcal{L}(e_n)\notag\\
=&\textbf{u}_{n-1}+e_n^*\textbf{z}_c(\textbf{x}_n)
\end{align}
and
\begin{align}
\label{12put2}
\textbf{v}_n=&\textbf{v}_{n-1}-\mu \nabla_{\textbf{v}} \mathcal{L}(e_n)\notag\\
=&\textbf{v}_{n-1}+\mu e_n^*\textbf{z}_c^*(\textbf{x}_n)
\end{align}
where $e_n=y_n-\textbf{u}^H_{n-1}\textbf{z}_c(\textbf{x}_n)-\textbf{v}^H_{n-1}\textbf{z}_c^*(\textbf{x}_n)$. The step-size $\mu$ controls the convergence rate of the proposed algorithm. Fig.~\ref{WLCRENF} illustrates the architecture of the WLRECF.

To simplify the notation, using an \textit{augmented} weight vector $\mathfrak{w}_n=[\textbf{u}^T_n~\textbf{v}^T_n]^T$ and a \textit{complex augmented} vector $\textbf{z}_{c,c}(\textbf{x}_n)=[\textbf{z}_{c}^T(\textbf{x}_n)~\textbf{z}_c^{*T}(\textbf{x}_n)]^T$, we can rewrite the
proposed WLRECF algorithm (\ref{12put1})-(\ref{12put2}) as
\begin{align}
\label{12pugdrt2}
\mathfrak{w}_n=\mathfrak{w}_{n-1}+\mu e_n^*\textbf{z}_{c,c}(\textbf{x}_n)
\end{align}

\section{Performance Analysis\label{sec:Per}}
In this section, the performances of the proposed schemes in terms of mean stability and
mean-square convergence are investigated. Instead of analyzing each proposed method separately,
we mainly focus on the WLRECF scheme that includes the LRECF method as a special case. In order to make the performance analysis tractable, some
assumptions are introduced.

\begin{assumption}\label{as2s1}
Inspired by (\ref{1235f24oput}), we consider an alternative observation model
\begin{align}
\label{12pug234drt2}
{y}_n=&\textbf{u}^H_{\textrm{opt},n}\textbf{z}_c(\textbf{x}_n)+\textbf{v}^H_{\textrm{opt},n}\textbf{z}_c^*(\textbf{x}_n)+\upsilon_n\notag\\
\triangleq&\mathfrak{w}^H_{\textrm{opt},n}\textbf{z}_{c,c}(\textbf{x}_n)+\upsilon_n
\end{align}
where $\mathfrak{w}_{\textrm{opt},n}=[\textbf{u}^T_{\textrm{opt},n}~\textbf{v}^T_{\textrm{opt},n}]^T$ represents the optimal augmented weight vector of an unknown system.
\end{assumption}

\begin{assumption}\label{as3s1}
In (\ref{12pug234drt2}), the noise $\upsilon_n$ is an independent and identically distributed (\textit{i.i.d.}) complex-valued random sequence with zero-mean and $E\{|\upsilon_n|^2\}=\sigma_{\upsilon}^2$, and is independent of the input $\mathbf{x}_j$ for all $j$.
\end{assumption}

\begin{assumption}\label{a2ss2_in}
In (\ref{12pug234drt2}), the time-varying unknown weight vector $\mathfrak{w}_{\textrm{opt},n}$ is defined as a random walk model, i.e.,
\begin{align*}
\mathfrak{w}_{\textrm{opt},n}=\mathfrak{w}_{\textrm{opt},n-1}+\mathbf{q}_n,
\end{align*}
where the random perturbation $\textbf{q}_n$ is a stationary white noise vector with zero mean and
covariance matrix $E\{\mathbf{q}_n\mathbf{q}^H_n\}=\sigma_q^2\mathbf{I}$, which is mutually independent of the input $\{\textbf{z}_c(\textbf{x}_n)\}$ and noise $\{\upsilon_n\}$.
\end{assumption}
\subsection{Mean Convergence Analysis}

Let the weight error vector
$
\tilde{\mathfrak{w}}_{n}=\mathfrak{w}_{\textrm{opt},n}-\mathfrak{w}_{n}
$.
The output error $e_n$ becomes
\begin{align}
e_n=\tilde{\mathfrak{w}}^H_{n-1}\textbf{z}_{c,c}(\textbf{x}_n)+\upsilon_n
\end{align}
while its conjugate is
\begin{align}
 \label{eq45uyi87}
e_n^*=\tilde{\mathfrak{w}}^T_{n-1}\textbf{z}_{c,c}^*(\textbf{x}_n)+\upsilon_n^*
\end{align}

Inserting (\ref{eq45uyi87}) into (\ref{12pugdrt2}), the recursion of the weight error vector $\tilde{\mathfrak{w}}_{n}$ is
\begin{align}
 \label{eq4st4687}
\tilde{\mathfrak{w}}_n=\tilde{\mathfrak{w}}_{n-1}-\mu (\tilde{\mathfrak{w}}^T_{n-1}\textbf{z}_{c,c}^*(\textbf{x}_n)+\upsilon_n^*)\textbf{z}_{c,c}(\textbf{x}_n)+\textbf{q}_n\notag\\
=\left(\mathbf{I}-\mu \textbf{z}_{c,c}(\textbf{x}_n)\textbf{z}_{c,c}^H(\textbf{x}_n)\right)\tilde{\mathfrak{w}}_{n-1}-\mu\upsilon_n^*\textbf{z}_{c,c}(\textbf{x}_n)+\textbf{q}_n
\end{align}

Using assumptions \ref{as3s1}-\ref{a2ss2_in} and well-known independence assumption \cite{IEEEhowto24,IEEEhowto14add,IEEEhowto25,IEEEhowto25add4},
we have $E\{\upsilon_n^*\textbf{z}_{c,c}(\textbf{x}_n)\}=\mathbf{0}$, $E\{\textbf{q}_n\}=\mathbf{0}$, and hence
\begin{align}
 \label{eq4s452t4687}
E\{\tilde{\mathfrak{w}}_n\}=\left(\mathbf{I}-\mu E\left\{\textbf{z}_{c,c}(\textbf{x}_n)\textbf{z}_{c,c}^H(\textbf{x}_n)\right\}\right)E\{\tilde{\mathfrak{w}}_{n-1}\}
\end{align}

Thus, when the step-size satisfies
\begin{align}
 \label{eq45fd4st4687}
0<\mu<\frac{2}{\lambda_{\max}\left(E\left\{\textbf{z}_{c,c}(\textbf{x}_n)\textbf{z}_{c,c}^H(\textbf{x}_n)\right\}\right)}
\end{align}
the proposed scheme is stable in the mean sense, and is unbiased, i.e., $E\{\tilde{\mathfrak{w}}_n\}\rightarrow\mathbf{0}$.
\subsection{Mean-Square Convergence Analysis}
The MSE performance is defined as
\begin{align}
 \label{eq5w4v467862}
E\{|e_n|^2\}=&E\{\tilde{\mathfrak{w}}^H_{n-1}\textbf{z}_{c,c}(\textbf{x}_n)\textbf{z}_{c,c}^H(\textbf{x}_n)\tilde{\mathfrak{w}}_{n-1}\}+\sigma_{\upsilon}^2
\notag\\
=&\Tr\{E\{\textbf{z}_{c,c}(\textbf{x}_n)\textbf{z}_{c,c}^H(\textbf{x}_n)\}E\{\tilde{\mathfrak{w}}_{n-1}\tilde{\mathfrak{w}}^H_{n-1}\}\}+\sigma_{\upsilon}^2\notag\\
\triangleq&\Tr\{\mathbf{R_z} E\{\tilde{\mathfrak{w}}_{n-1}\tilde{\mathfrak{w}}^H_{n-1}\}\}+\sigma_{\upsilon}^2
\end{align}
where $\mathbf{R_z}=\{\textbf{z}_{c,c}(\textbf{x}_n)\textbf{z}_{c,c}^H(\textbf{x}_n)\}$.

Upon multiplying both sides of (\ref{eq4st4687}) by $\tilde{\mathfrak{w}}^H_{n}$ yields the following relation
\begin{align}
 \label{eq5467862}
&\veq \tilde{\mathfrak{w}}_{n}\tilde{\mathfrak{w}}^H_{n}\notag\\
&=\tilde{\mathfrak{w}}_{n-1}\tilde{\mathfrak{w}}^H_{n-1}-\mu \textbf{z}_{c,c}(\textbf{x}_n)\textbf{z}_{c,c}^H(\textbf{x}_n)\tilde{\mathfrak{w}}_{n-1}\tilde{\mathfrak{w}}^H_{n-1}\notag\\
&\veq -\mu \tilde{\mathfrak{w}}_{n-1}\tilde{\mathfrak{w}}^H_{n-1}\textbf{z}_{c,c}(\textbf{x}_n)\textbf{z}_{c,c}^H(\textbf{x}_n)\notag\\
&\veq +\mu^2\textbf{z}_{c,c}(\textbf{x}_n)\textbf{z}_{c,c}^H(\textbf{x}_n)\tilde{\mathfrak{w}}_{n-1}\tilde{\mathfrak{w}}^H_{n-1}\textbf{z}_{c,c}(\textbf{x}_n)\textbf{z}_{c,c}^H(\textbf{x}_n)\notag\\
&\veq +\mu^2|\upsilon_n|^2\textbf{z}_{c,c}(\textbf{x}_n)\textbf{z}_{c,c}^H(\textbf{x}_n) +\mathbf{q}_n\mathbf{q}^H_n\notag\\
&\veq -\mu\upsilon_n\left(\mathbf{I}-\mu \textbf{z}_{c,c}(\textbf{x}_n)\textbf{z}_{c,c}^H(\textbf{x}_n)\right)\tilde{\mathfrak{w}}_{n-1}\textbf{z}_{c,c}^H(\textbf{x}_n)\notag\\
&\veq -\mu\upsilon_n^*\textbf{z}_{c,c}(\textbf{x}_n)\tilde{\mathfrak{w}}^H_{n-1}\left(\mathbf{I}-\mu \textbf{z}_{c,c}(\textbf{x}_n)\textbf{z}_{c,c}^H(\textbf{x}_n)\right)\notag\\
&\veq + \left(\mathbf{I}-\mu \textbf{z}_{c,c}(\textbf{x}_n)\textbf{z}_{c,c}^H(\textbf{x}_n)\right)\tilde{\mathfrak{w}}_{n-1}\mathbf{q}^H_n\notag\\
&\veq + \mathbf{q}_n\tilde{\mathfrak{w}}^H_{n-1}\left(\mathbf{I}-\mu \textbf{z}_{c,c}(\textbf{x}_n)\textbf{z}_{c,c}^H(\textbf{x}_n)\right)\notag\\
&\veq -\mu\upsilon_n^*\textbf{z}_{c,c}(\textbf{x}_n) \mathbf{q}_n^H -\mu\upsilon_n\mathbf{q}_n\textbf{z}_{c,c}^H(\textbf{x}_n).
\end{align}

Proceeding in a manner similar to \cite{IEEEhowto21,IEEEhowto22} and using assumptions \ref{as3s1}-\ref{a2ss2_in}\footnote{Under the assumptions \ref{as3s1}-\ref{a2ss2_in}, we know that the last six terms in (\ref{eq5467862}) are equal to zero.}
\begin{align}
 \label{eq454687}
&\vect({E}\{\tilde{\mathfrak{w}}_{n}\tilde{\mathfrak{w}}^H_{n}\})\notag\\
=&(\mathbf{I}-\mu \mathbf{A}+\mu^2\mathbf{B})\vect(E\{\tilde{\mathfrak{w}}_{n-1}\tilde{\mathfrak{w}}^H_{n-1}\})\notag\\
&+\mu^2\sigma_{\upsilon}^2\vect(\mathbf{R_z})+\sigma_q^2\vect(\mathbf{I})
\end{align}
where
\begin{align}
\label{e15tyrw646}
\mathbf{A}&=\mathbf{I}\otimes E\{\textbf{z}_{c,c}\textbf{x}_n)\textbf{z}_{c,c}^H(\textbf{x}_n)\}+E\{\textbf{z}_{c,c}^*(\textbf{x}_n)\textbf{z}_{c,c}^T(\textbf{x}_n)\}\otimes \mathbf{I},\\
\label{e15tyrw646_2}
\mathbf{B}&=E\{(\textbf{z}_{c,c}^*(\textbf{x}_n)\textbf{z}_{c,c}^T(\textbf{x}_n))\otimes(\textbf{z}_{c,c}(\textbf{x}_n)\textbf{z}_{c,c}^H(\textbf{x}_n))\},
\end{align}

Thus, when the step-size is such that the matrix $\mathbf{I}-\mu \mathbf{A}+\mu^2\mathbf{B}$ is stable, the proposed algorithm is convergent in the mean-square sense. Moreover, we can obtain the steady-state mean-square deviation (MSD) 
\begin{align}
 \label{eq454687}
  \lim_{i\rightarrow\infty}E\{\tilde{\mathfrak{w}}_{n}\tilde{\mathfrak{w}}^H_{n}\}
=&\mu\sigma_{\upsilon}^2\vect^{-1}\{( \mathbf{A}-\mu\mathbf{B})^{-1}\vect(\mathbf{R_z})\}\notag\\
&+\frac{\sigma_q^2\vect^{-1}\{(\mathbf{A}-\mu\mathbf{B})^{-1}\vect(\mathbf{I})\}}{\mu}
\end{align}

Inserting (\ref{eq454687}) into (\ref{eq5w4v467862}), we finally get the steady-state MSE
\begin{align}
 \label{eq56sjt4532}
  {\rm MSE_{nonsta}}
=&\sigma_{\upsilon}^2+\mu\sigma_{\upsilon}^2\vect^T(\mathbf{R_z})( \mathbf{A}-\mu\mathbf{B})^{-1}\vect(\mathbf{R_z})\notag\\
&+\frac{\sigma_q^2\vect^T(\mathbf{R_z})(\mathbf{A}-\mu\mathbf{B})^{-1}\vect(\mathbf{I})}{\mu}
\end{align}


Remark: (i) A sufficiently small step-size can guarantee the proposed algorithm to be stable. This is because when $\mu$ is sufficiently small, the terms $\mu\mathbf{B}$ compared with $\mathbf{A}$ can be neglected, i.e., $\mathbf{I}-\mu \mathbf{A}+\mu^2\mathbf{B}\approx\mathbf{I}-\mu \mathbf{A}$. In this case, the step-size should
satisfy $0<\mu<\frac{2}{\lambda_{\max}\left(\mathbf{A}\right)}$.

(ii) For a stationary system ($\mathbf{q}(n)=\mathbf{0}$), (\ref{eq56sjt4532}) simplifies to ${\rm MSE_{sta}}
=\sigma_{\upsilon}^2+\mu\sigma_{\upsilon}^2\vect^T(\mathbf{R_z})( \mathbf{A}-\mu\mathbf{B})^{-1}\vect(\mathbf{R_z})
$. When the used step-size $\mu\rightarrow0$, ${\rm MSE_{sta}}$ tends to the minimum $\sigma_{\upsilon}^2$.

(iii) For the non-stationary system ($\mathbf{q}(n)\neq\mathbf{0}$), we know that there is an optimum step-size given by
$\mu_{\rm opt}=\frac{\sigma_q}{\sigma_{\upsilon}}
\sqrt{\frac{\phi}{\varphi}}$,
and the corresponding minimum MSE is
$  {\rm MSE_{nonsta,min}}
=\sigma_{\upsilon}^2+2\sigma_{\upsilon}\sigma_q\sqrt{\varphi\phi}$,
with $\phi=\vect^T(\mathbf{R_z})\mathbf{A}^{-1}\vect(\mathbf{I}),~\varphi=\vect^T(\mathbf{R_z})\mathbf{A}^{-1}\vect(\mathbf{R_z})$.
\begin{figure*}[!htp]
\centering
\hspace*{-1em}
\subfigure[]{\includegraphics[height=58mm]{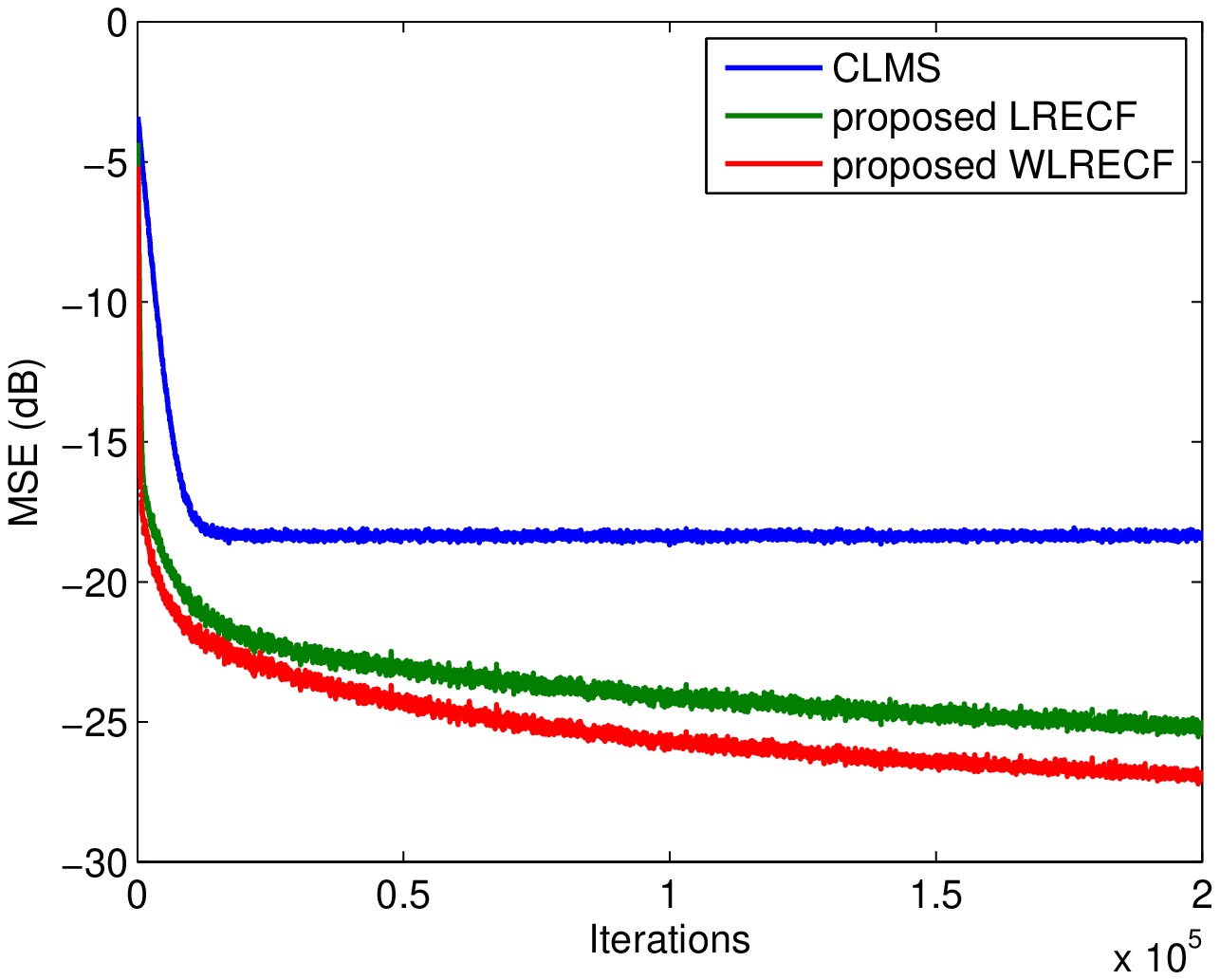}}
\subfigure[]{\includegraphics[height=58mm]{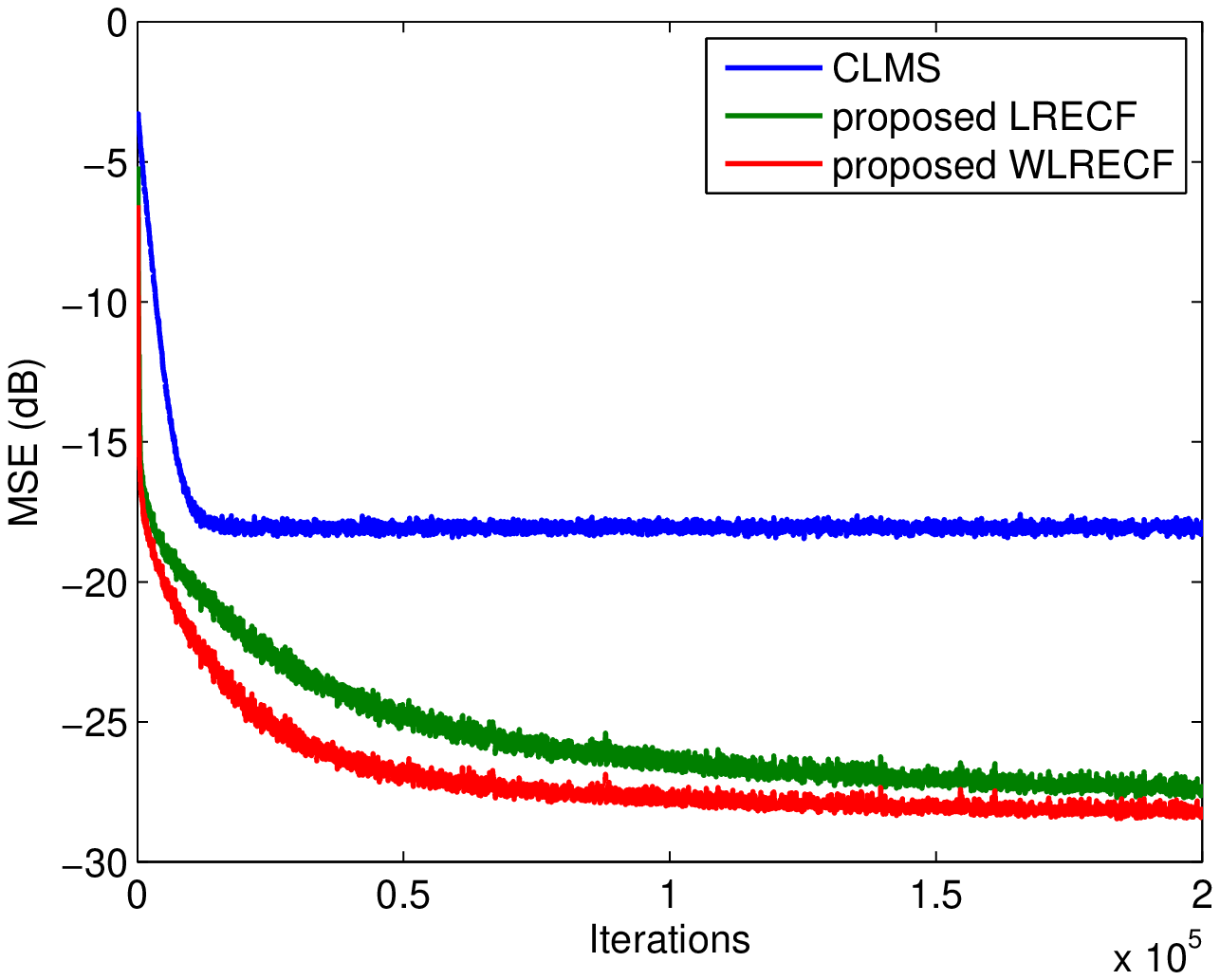}}
\vspace{-1ex}
\caption{Performance comparison between the CLMS, LRECF, and WLRECF for the nonlinear system I. (a) the circular
input, (b) the noncircular input $\rho=0.1$.}
\label{sim_fig1}
\end{figure*}
\begin{figure*}[!t]
\centering
\hspace*{-1em}
\subfigure[]{\includegraphics[height=58mm]{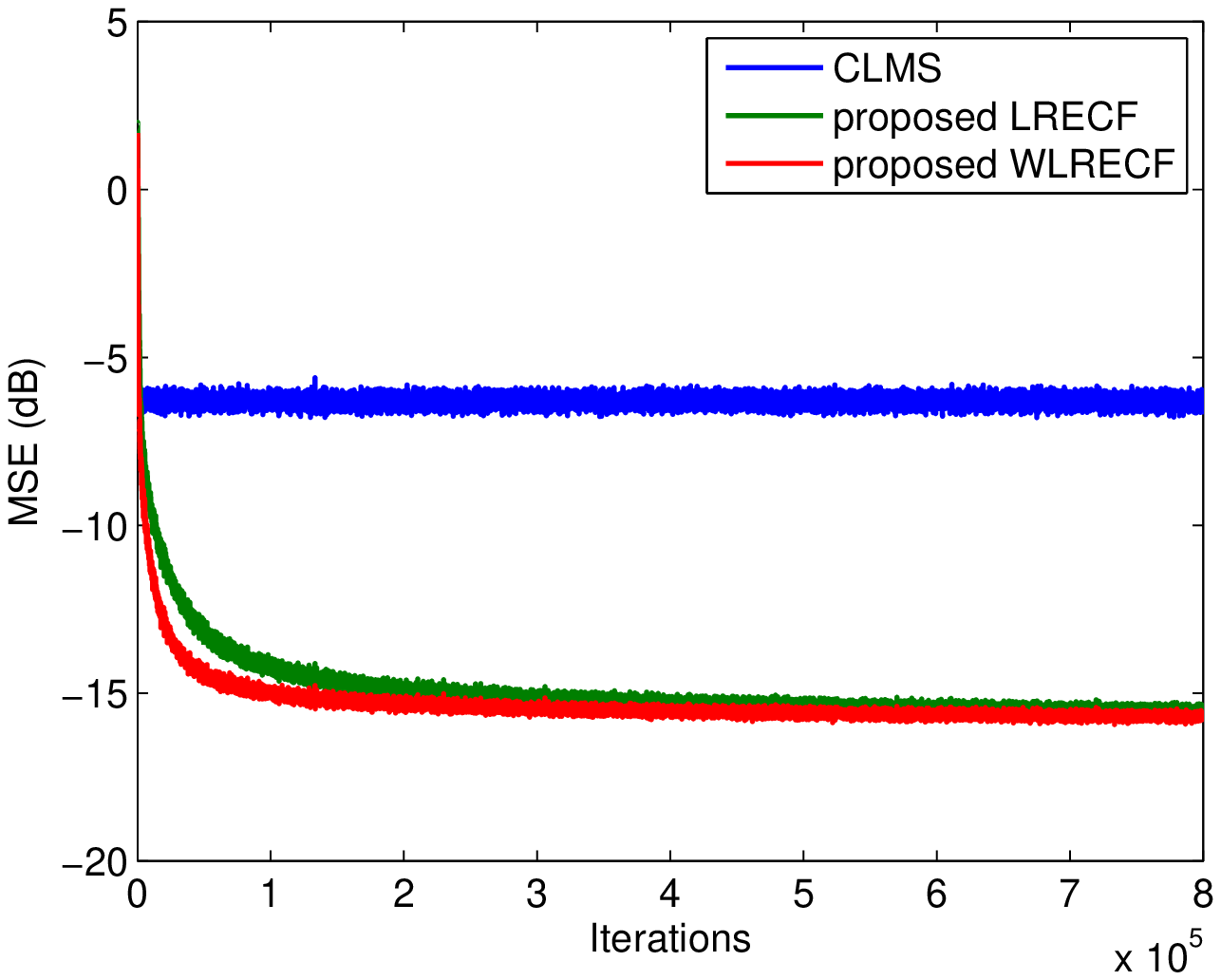}}
\subfigure[]{\includegraphics[height=58mm]{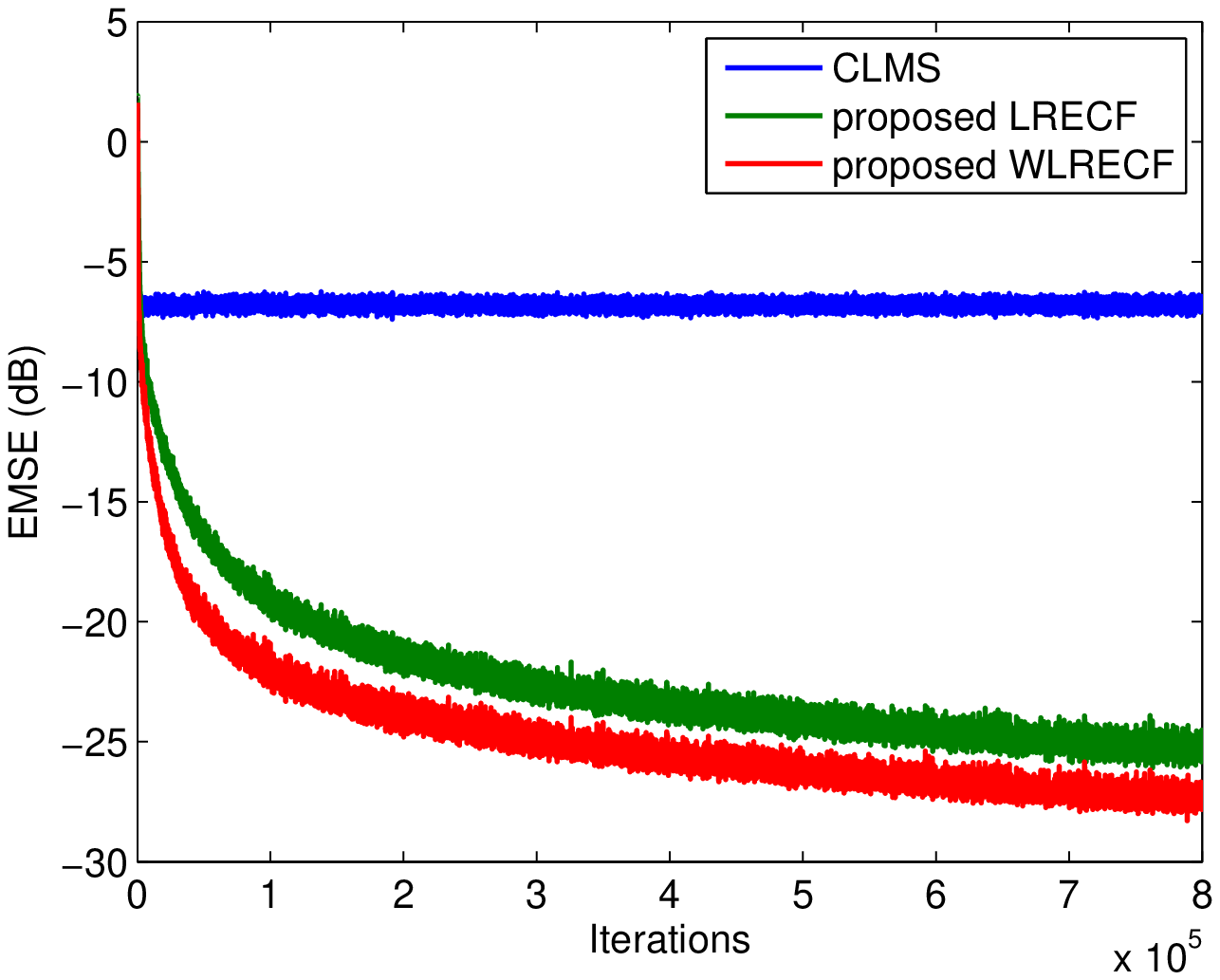}}
\vspace{-1ex}
\caption{Performance comparison between the CLMS, LRECF, and WLRECF for the nonlinear system II. (a) the MSE curves, (b) the EMSE curves.}
\label{sim_fig2}
\end{figure*}
\section{Monte Carlo Simulations\label{sec:MON}}
In this section, Monte Carlo simulations are presented. First, to examine the convergence performance of the proposed two filters, the nonlinear system identification task is carried out. Then, a nonlinear channel equalization task is considered.
To evaluate the filtering performance, the MSE in dB is used and defined as
\begin{align*}
{\rm MSE} = 10{\rm log}_{10}(E\{|e_n|^2\})
\end{align*}
where the expectation is obtained by averaging the results of $200$ independent runs.
\subsection{Nonlinear system identification}
In the complex-valued nonlinear system identification, we consider two different nonlinear systems used in \cite{IEEEFAD2012,IEEEPKS2012}. Experiments are conducted on a set of the input-output signal $\{x_{n},x_{n-1},\cdots,x_{n-m+1},y_{n}\}$ with $m>0$.
\subsubsection{Nonlinear System I} The first nonlinear system is chosen as
\begin{align}
 \label{eq56sjt4gh}
y_n=t_n+(0.15-0.1i)t^2_n
\end{align}
where $t_n$ is an output signal of a linear filter
\begin{align}
 \label{eq5rt6sjt4532}
t_n=\sum_{k=1}^{5}h_kx_{n-k+1}
\end{align}
with $h_k$ being
\begin{align}\label{eq56sjgert4532}
h_k=
0.432\Big(1+\cos\Big(\frac{2\pi(k-3)}{5}\Big)\notag\\
-i\Big(1+\cos\Big(\frac{2\pi(k-3)}{10}\Big)\Big)\Big)
\end{align}
\begin{figure}[!t]
\centering
\includegraphics[height=58mm]{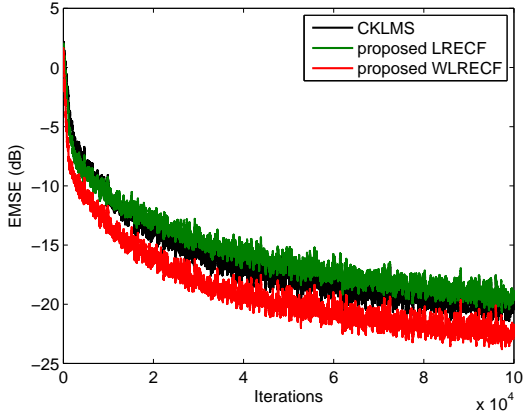}
\vspace{-1ex}
\caption{The EMSE comparison between the CKLMS, LRECF, and WLRECF.}
\label{simf}
\end{figure}
\begin{figure*}[!t]
\centering
\hspace*{-1em}
\subfigure[]{\includegraphics[height=45mm]{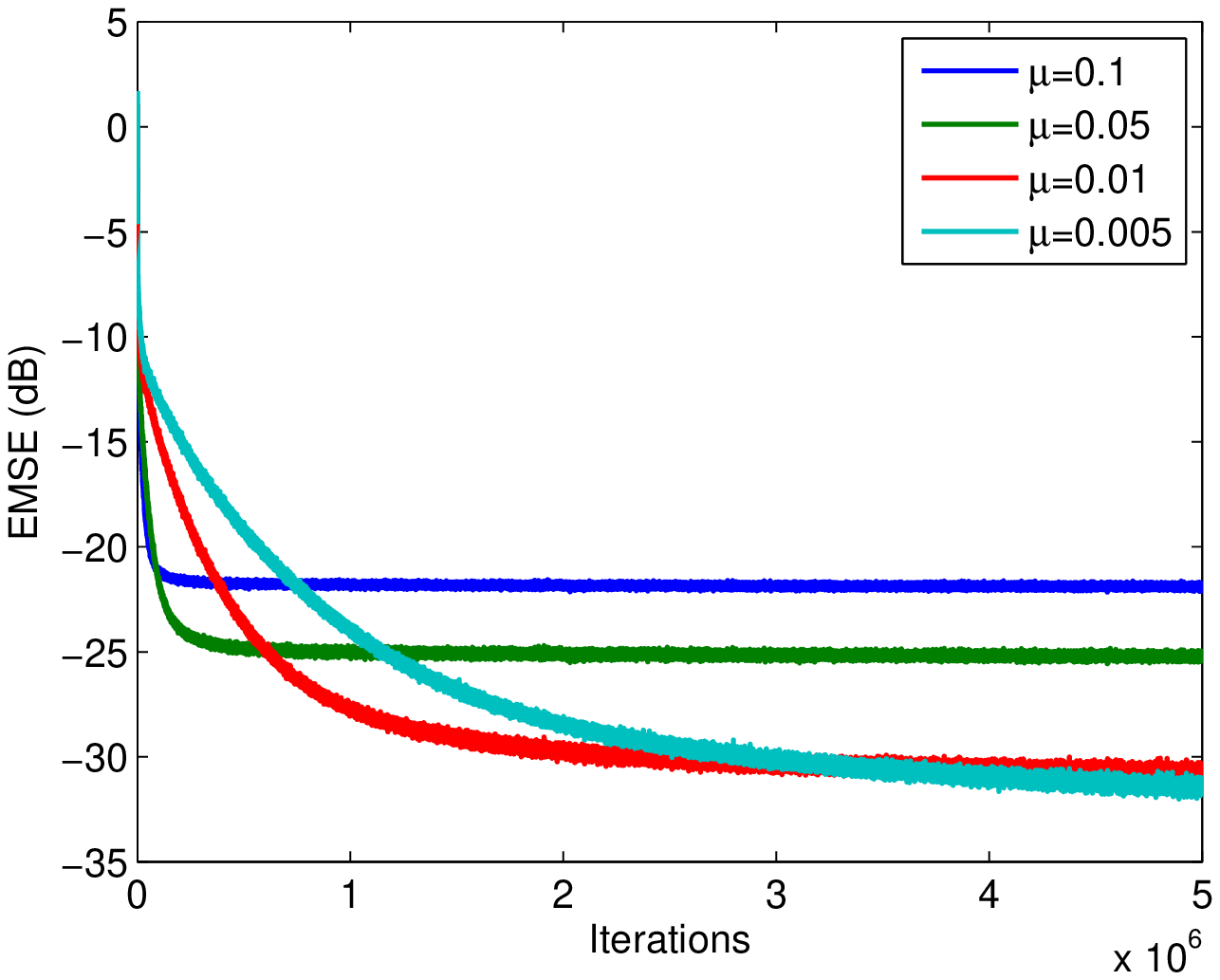}}
\subfigure[]{\includegraphics[height=45mm]{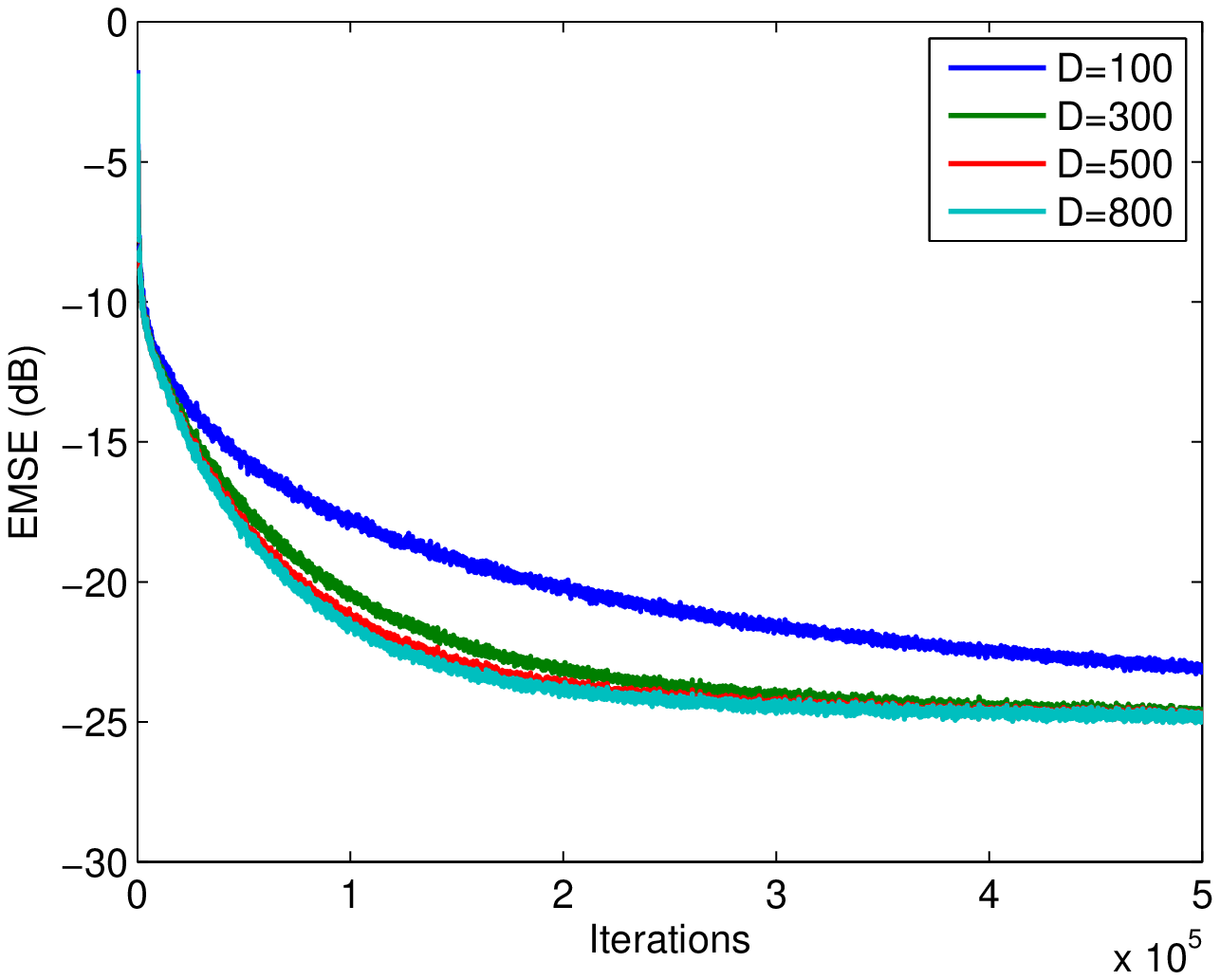}}
\subfigure[]{\includegraphics[height=45mm]{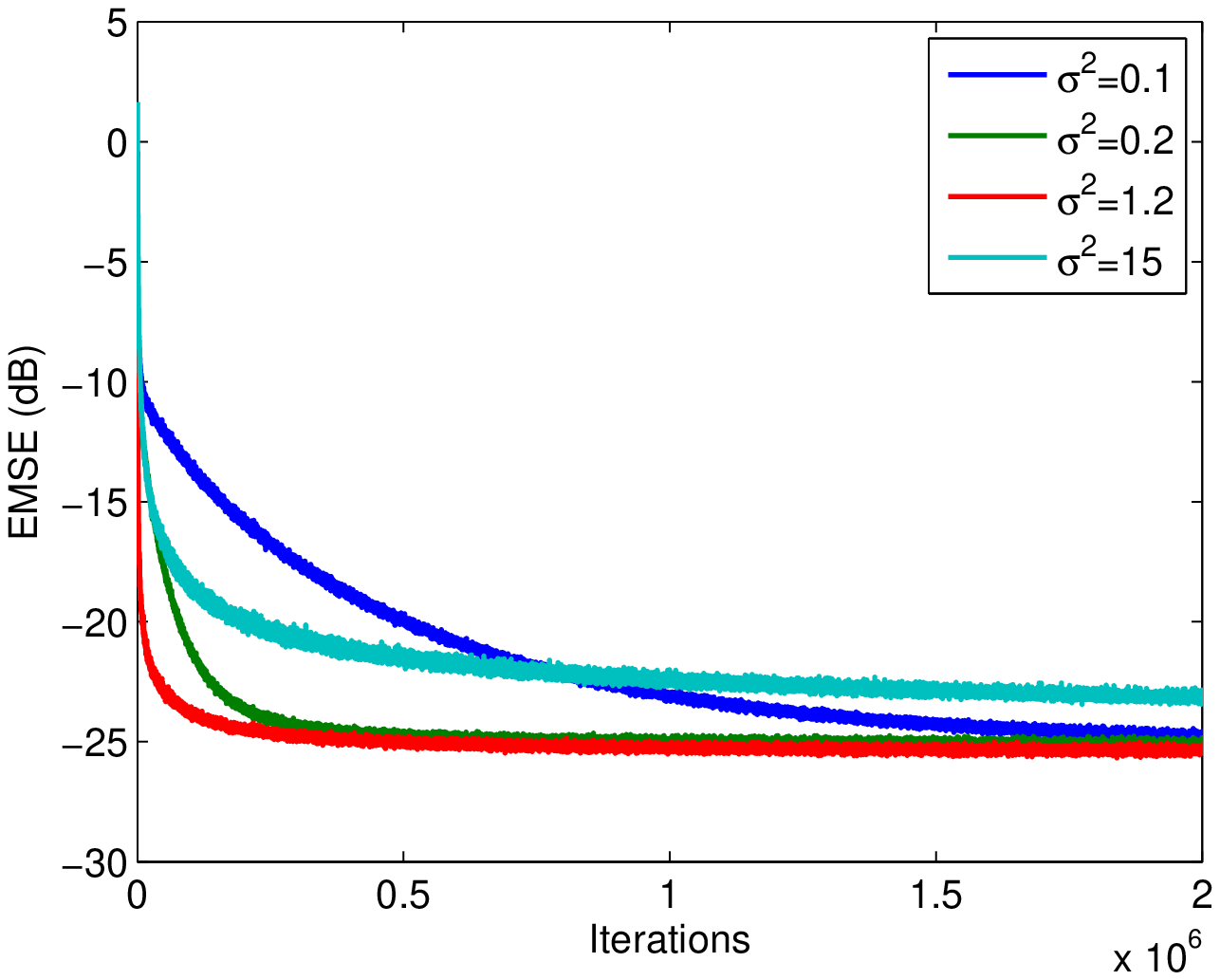}}
\vspace{-1ex}
\caption{EMSE curves of the proposed WLRECF with different step-sizes, $D$, and $\sigma^2$. (a) Effect of $\mu$, (b) Effect of $D$, (c) Effect of $\sigma^2$.}
\label{sim_fig2ddd}
\end{figure*}
\begin{table}
\centering
  \caption{Comparison of Averaged Consumed Time Over $1\times10^{5}$ Learning Samples}
  \begin{tabular}{cc}
    \hline
    \hline
    Adaptive filters      & Averaged consumed time ($s$)\\
     \hline
    CKLMS  & $7.83\times10^{3}$ \\
    LRECF  & $21.61$ \\
    WLRECF & $52.3$ \\
    \hline
    \hline
 \end{tabular}
\label{tabld}
\end{table}

In the input-output relationship (\ref{eq56sjt4gh})-(\ref{eq56sjgert4532}), $y_n$ and $x_n$ are complex-valued output and input signals. At the receiver end of the system, the output signal is corrupted by white Gaussian noise and the level of the noise is set to 30 dB. The input signal is obtained by the using of the form
\begin{align}\label{esjgert4532}
x_n=\sqrt{1-\rho^2}s_{_{1,n}}+i\rho s_{_{2,n}}
\end{align}
where $s_{_{1,n}}$, $s_{_{2,n}}$ are zero-mean Gaussian random variables, $\rho\in[0,1]$ determines the performance of $x_n$. If $\rho$ approaches 0 or 1, the input is highly noncircular and for $\rho=\frac{\sqrt{2}}{2}$ it is circular. The step-size of the complex-valued least mean square (CLMS) is $\mu=0.05$. For a fair comparison, the step-sizes of proposed algorithms are also set to $\mu=0.05$. The other parameters are $m=5, D=500, \sigma^2=0.2$.
In Fig.~\ref{sim_fig1}, the MSE learning curves for circular and noncircular input signals are depicted.
\subsubsection{Nonlinear System II} The second nonlinear system consists of
a linear filter:
\begin{align}
t_n=(-0.9+0.8i)x_n+(0.6-0.7i)x_{n-1}
\end{align}
and a memoryless nonlinearity
\begin{align}
y_n=t_n+(0.1+0.15i)t^2_n+(0.06+0.05i)t^3_n
\end{align}

The input signal has the form $x_n=x_{_{1,n}}+ix_{_{2,n}}$, where $x_{_{1,n}},~x_{_{2,n}}$ are uniform randomly distributed signals and their ranges are $[-1,1]$. The observation noise corrupts the output signal with the variance 16 dB. The step-sizes of the CLMS, LRECF, and WLRECF are $0.005$. The other parameters are set to be $m=2, D=500, \sigma^2=1$. 
The MSE and excess mean-square error (EMSE) learning curves are shown in Fig.~\ref{sim_fig2}, where the EMSE is defined as
${\rm EMSE}=10{\rm log}_{10}(E\{|y_n-\hat{y}_n-\upsilon_n|^2\})$ with $\upsilon_n$ being the noise added to the desired signal.

Both Fig.~\ref{sim_fig1} and Fig.~\ref{sim_fig2} show that the proposed schemes can achieve an improved performance compared with the CLMS. It is also shown that in proposed two methods the LRECF is inferior to the WLRECF. From the Section~\ref{sec:Pro}, we know the LRECF is an approximation for the CKLMS. Due to the growing network, the CKLMS poses both computational as well as memory issues for large learning samples, such as $8\times10^5$ samples. Thus, we only compare the CKLMS with the proposed methods for relatively small samples drawn in Fig.~\ref{simf}, and the averaged consumed time is listed in Table~\ref{tabld}. It is measured on a 3.2
GHz Intel Core i5 processor with 8 Gb of RAM, running Matlab R2017a on Windows 10. The experiment settings are the same as those used in Fig.~\ref{sim_fig2}. It demonstrates that the LRECF effectively approximates the CKLMS with lower complexity. The proposed WLRECF method could outperform the CKLMS and LRECF schemes. But the WLRECF requires more computations than the LRECF.

\subsubsection{Effect of the step-size, $D$, and $\sigma^2$}
To examine the effect of the step-size, $D$, and $\sigma^2$ on the performance of the proposed schemes, the EMSE curves of the proposed WLRECF with different $\mu$, $D$, and $\sigma^2$ are displayed in Fig.~\ref{sim_fig2ddd}. In Fig.~\ref{sim_fig2ddd}(a), $D=500,~\sigma^2=0.2$. In Fig.~\ref{sim_fig2ddd}(b), $\mu=0.05,~\sigma^2=0.2$. In Fig.~\ref{sim_fig2ddd}(c), $D=500,~\mu=0.05$. The other experiment settings are the same as those used in Fig.~\ref{sim_fig2}. As can be seen, the choice of the step-size determines a compromise between fast convergence rate and small steady-state EMSE. With fixed $\mu$ and $\sigma^2$, small $D$ (i.e., $D=100$) will suffer from slow convergence rate. Large $D$ can lead to improved convergence performance but with high computational cost, shown in Fig.~\ref{sim_fig2ddd}(b). With fixed $\mu$ and $D$, too large and two small $\sigma^2$ will suffer from poor convergence performance illustrated by Fig.~\ref{sim_fig2ddd}(c). Hence, to achieve fast convergence rate and low steady-state error, $\mu,~D$ and $\sigma^2$ should be chosen appropriately according to the application.
\begin{figure*}[!t]
\centering
\subfigure[]{\includegraphics[height=32mm]{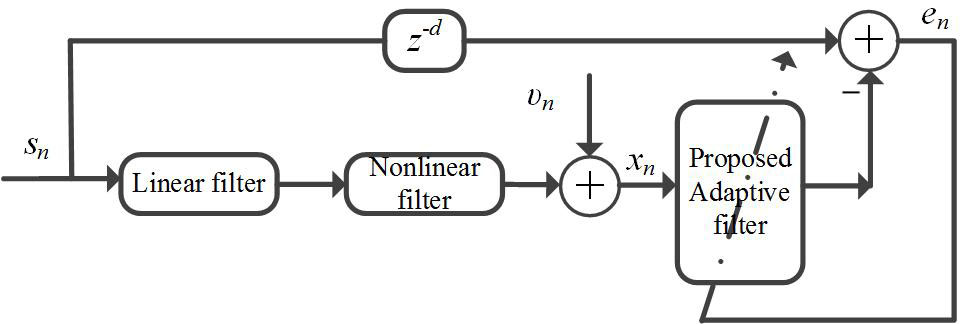}}\\
\vspace{-1ex}
\subfigure[]{\includegraphics[height=58mm]{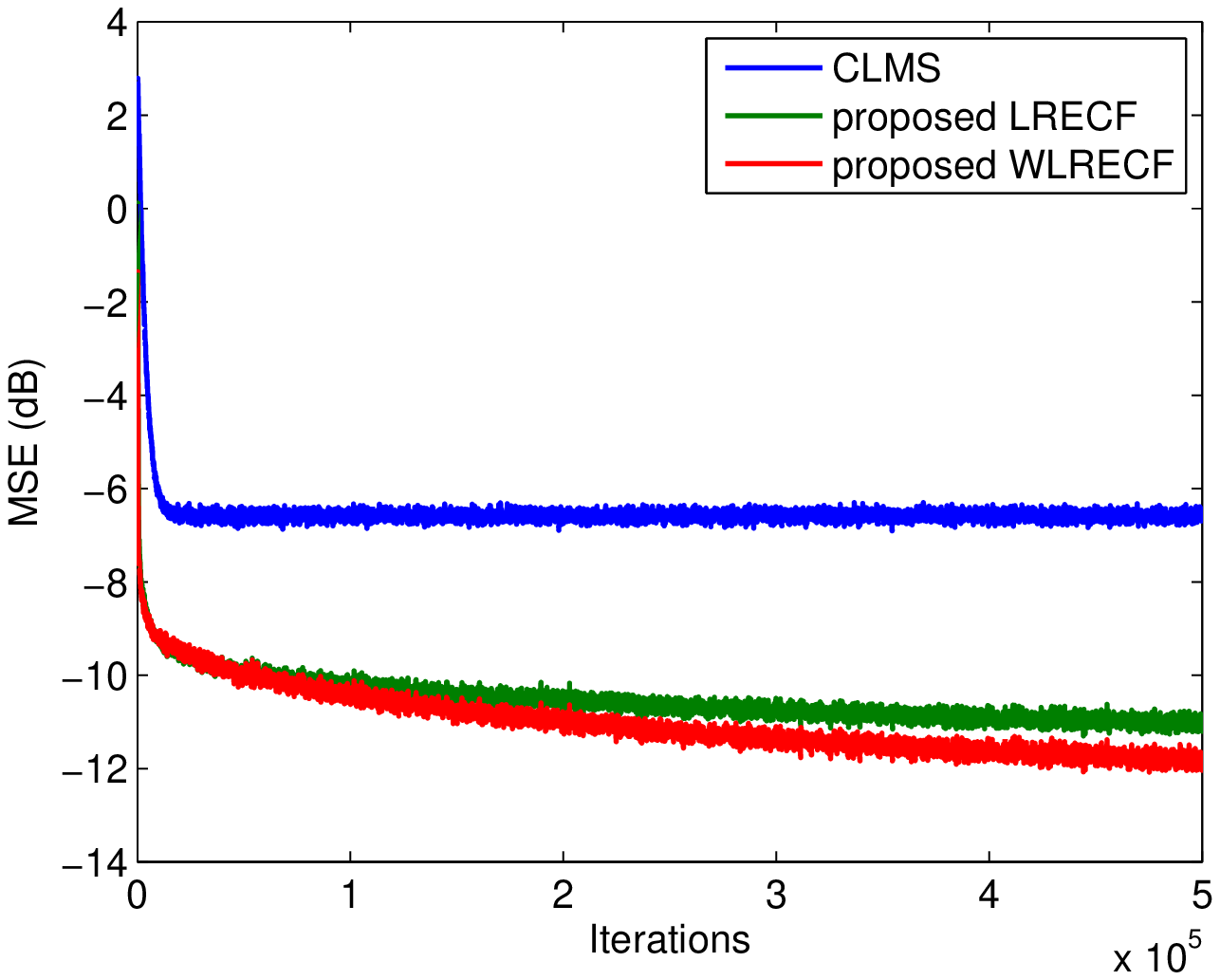}}
\subfigure[]{\includegraphics[height=58mm]{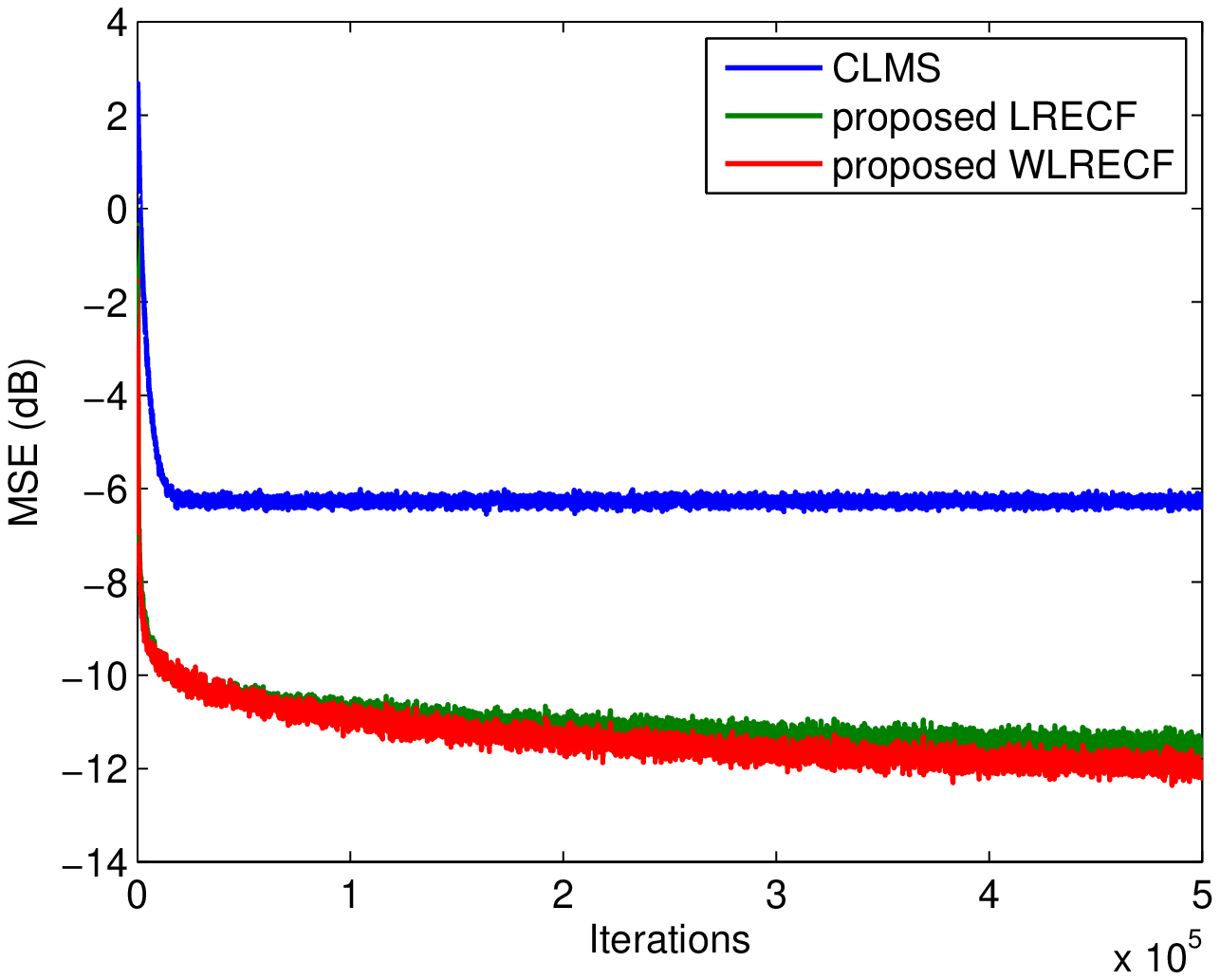}}
\vspace{-1ex}
\caption{The MSE comparison between the CLMS, CKLMS, LRECF, and WLRECF for the nonlinear channel equalization problem. (a) the nonlinear channel equalization, (b) the case 1, (c) the case 2.}
\label{sim_fig3}
\end{figure*}
\begin{figure*}[!t]
\centering
\hspace*{-1em}
\subfigure[]{\includegraphics[height=33.5mm]{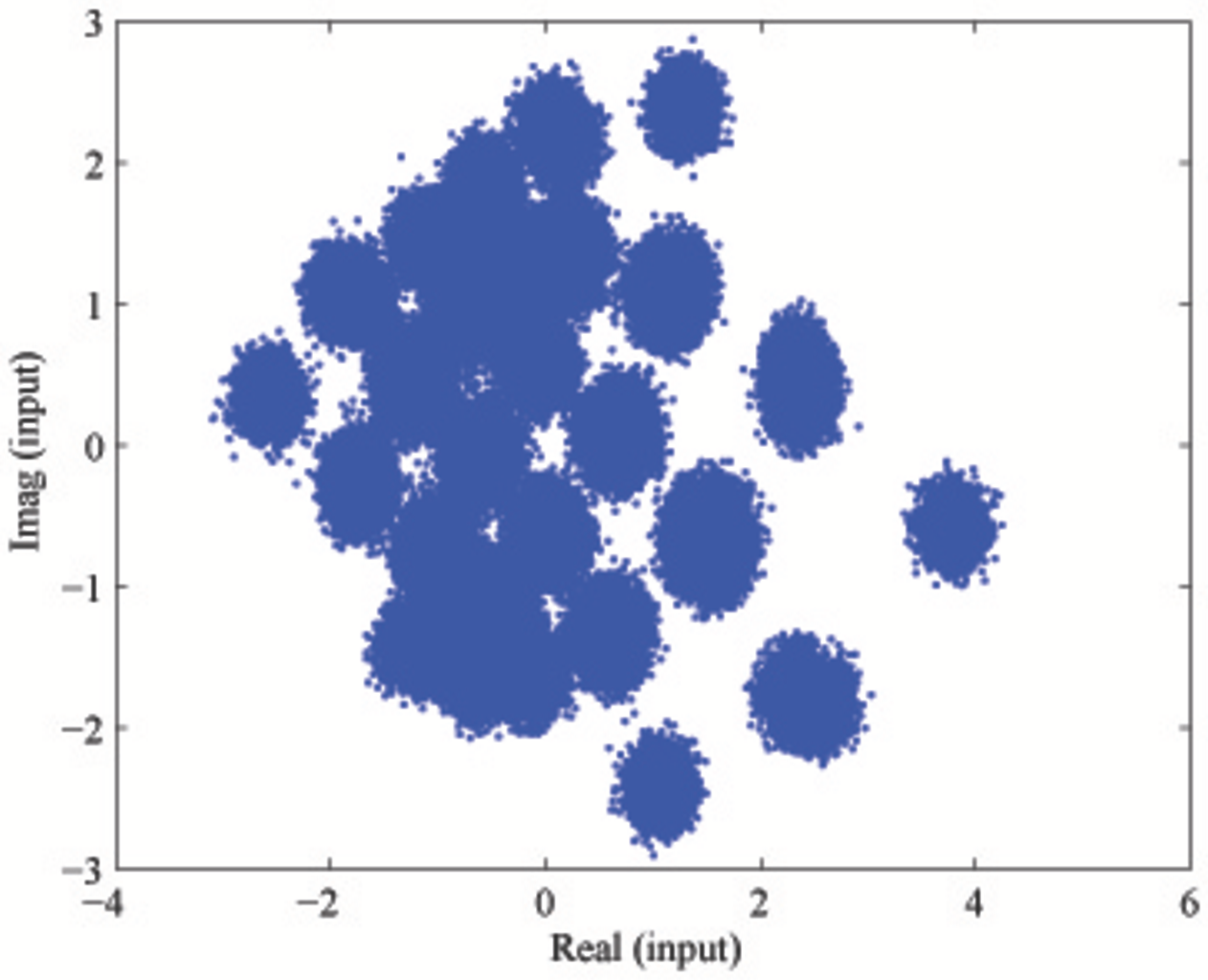}}
\subfigure[]{\includegraphics[height=33.5mm]{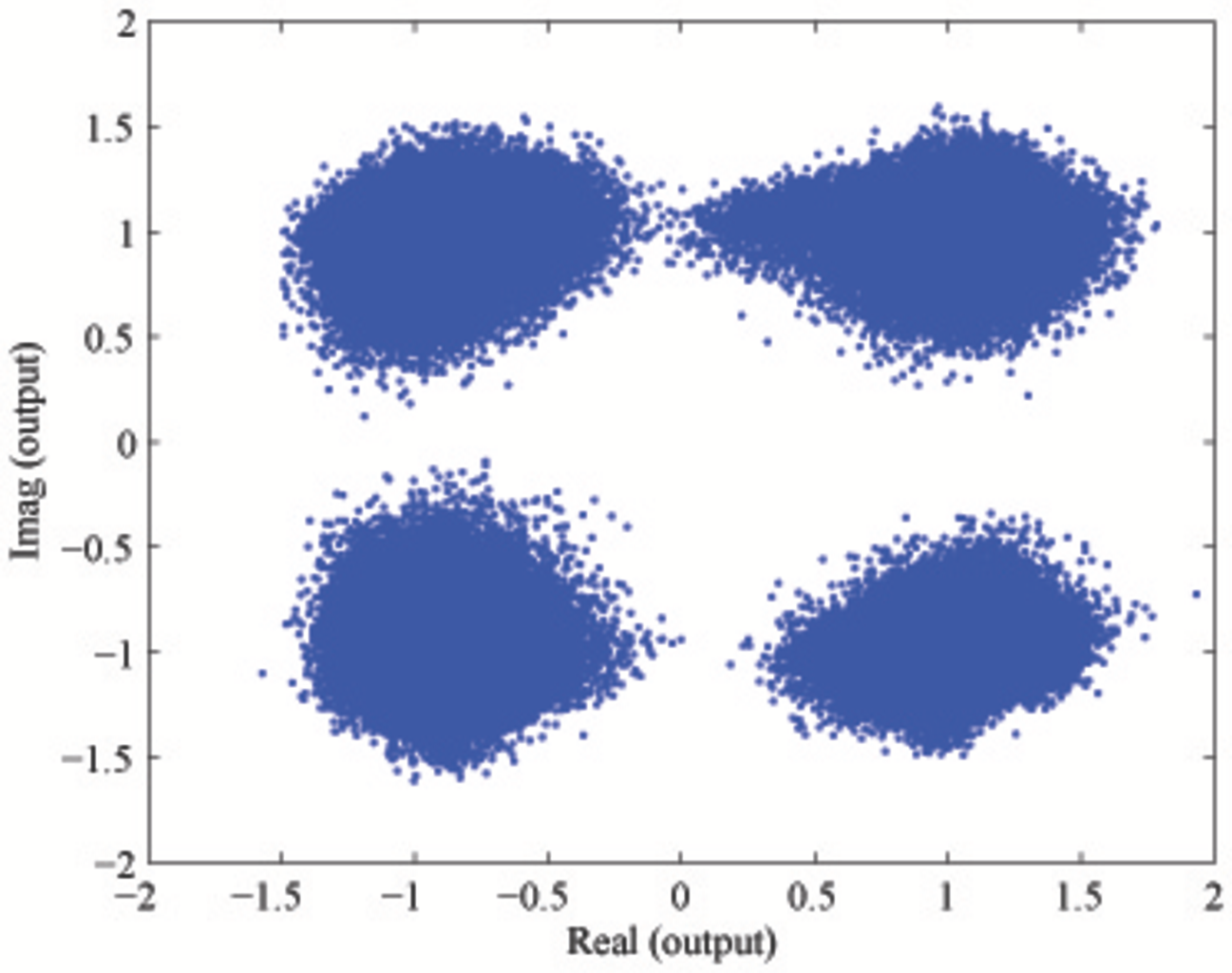}}
\subfigure[]{\includegraphics[height=33.5mm]{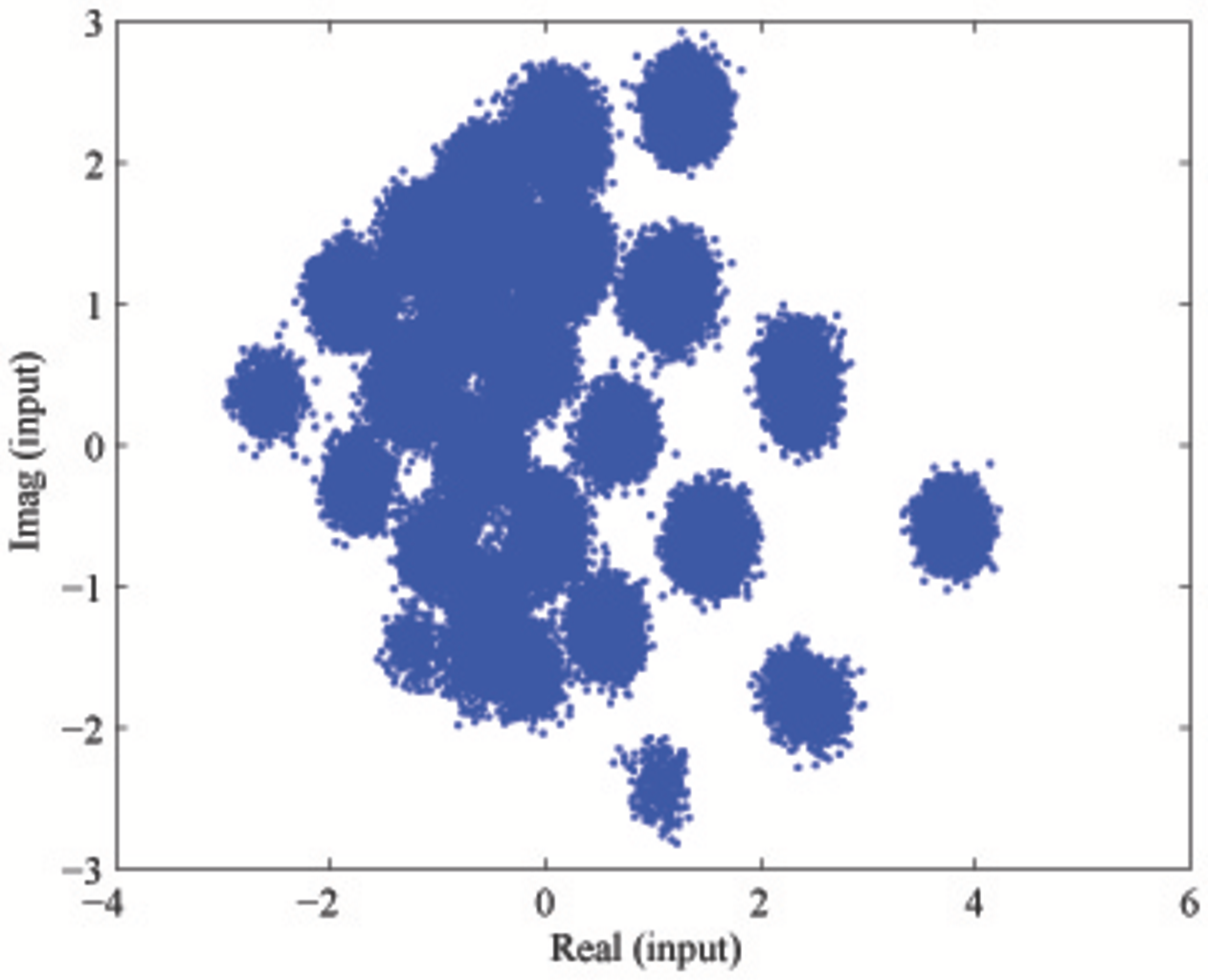}}
\subfigure[]{\includegraphics[height=33.5mm]{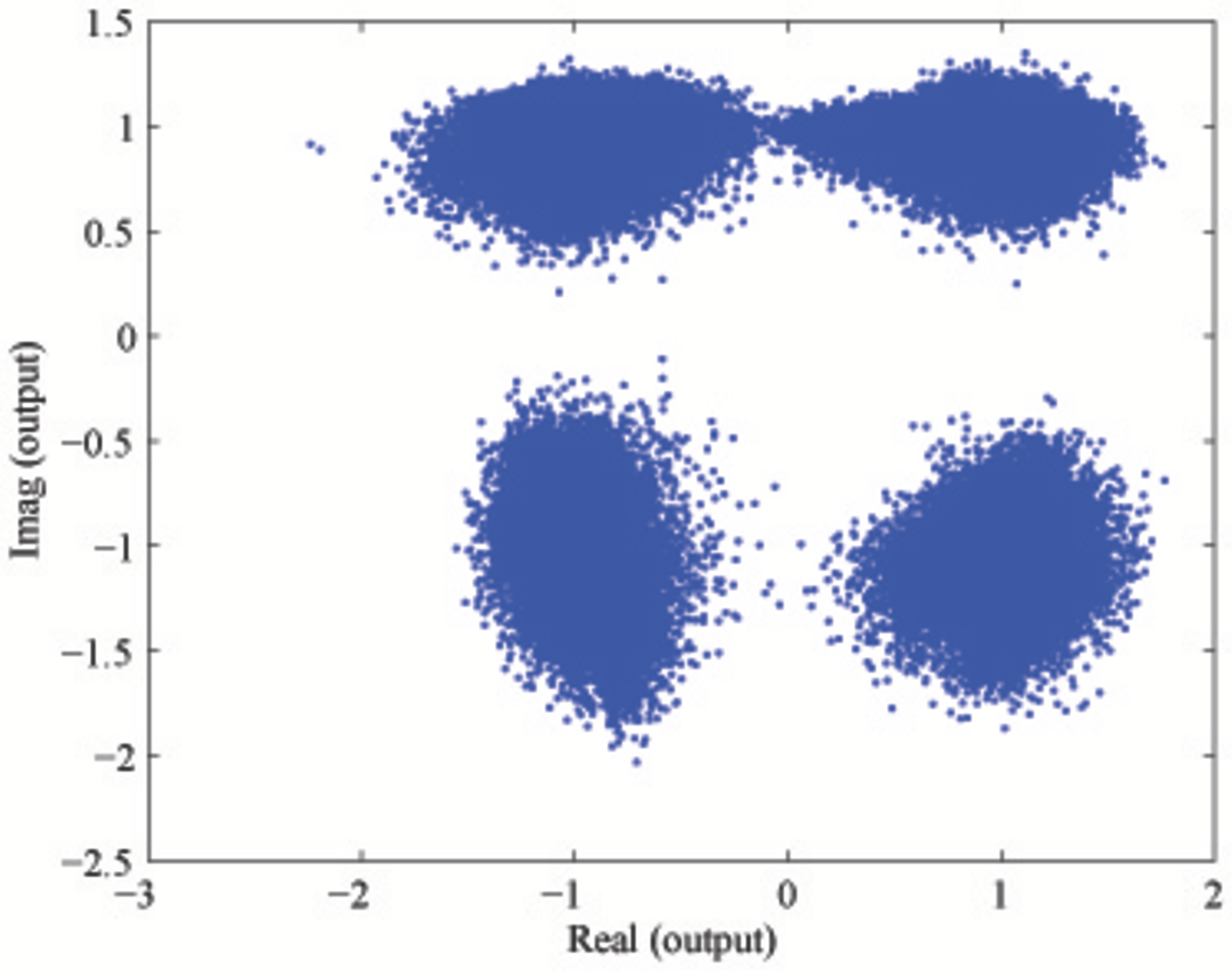}}
\vspace{-1ex}
\caption{Eye diagram of symbol classification performance using the proposed WLRECF. (a) the input signal $x_n$ for the case 1, (b) the output signal $\hat{y}_n$ for the case 1, (c) the output signal $x_n$ for the case 2, (d) the output signal $\hat{y}_n$ for the case 2.}
\label{sim_fig3_1}
\end{figure*}
\subsection{Nonlinear channel equalization}
In the nonlinear channel equalization scenarios, we considered the equalization model which consists of a linear filter \cite{IEEEISA1995,IEEEMGBPN2005}
\begin{align}
t_n=&(0.34-0.27i)s_n+(0.87+0.43i)s_{n-1}\notag\\
&+(0.34-0.21i)s_{n-2}
\end{align}
and a nonlinear distortion
\begin{align}
x_n=t_n+0.1t^2_n+0.05t^3_n
\end{align}

The nonlinear channel equalization structure is shown in Fig.~\ref{sim_fig3}(a). The 4 QPSK symbols, $s_1=1+j,~s_2=1-j,~s_3=-1+j,~s_4=-1-j$, are tested: The case 1) the 4 symbols are equiprobable; 
The case 2) the occurrence probability are $p_1=0.4,~p_2=0.1,~p_3=0.4,~p_4=0.1$. 

The learning curves using the set of the training samples $\{x_{n},x_{n-1},\cdots,x_{n-m+1},s_{n-d}\}$ are drawn in Fig.~\ref{sim_fig3}(b)-(c), where $m>0$ and $d$ is the equalization time delay. The additive observation noise is a zero-mean Gaussian signal with variance 15 dB. The values of the parameters are $m=5,D=500,\sigma^2=0.05,~d=2, \mu=0.08$. By the use of $300000$ testing samples, Fig.~\ref{sim_fig3_1} gives the symbol classification performance of the equalizers with the proposed WLRECF.

\subsection{Theoretical Curves}

\begin{figure*}[!t]
\centering
\hspace*{-1em}
\subfigure[]{\includegraphics[height=60mm]{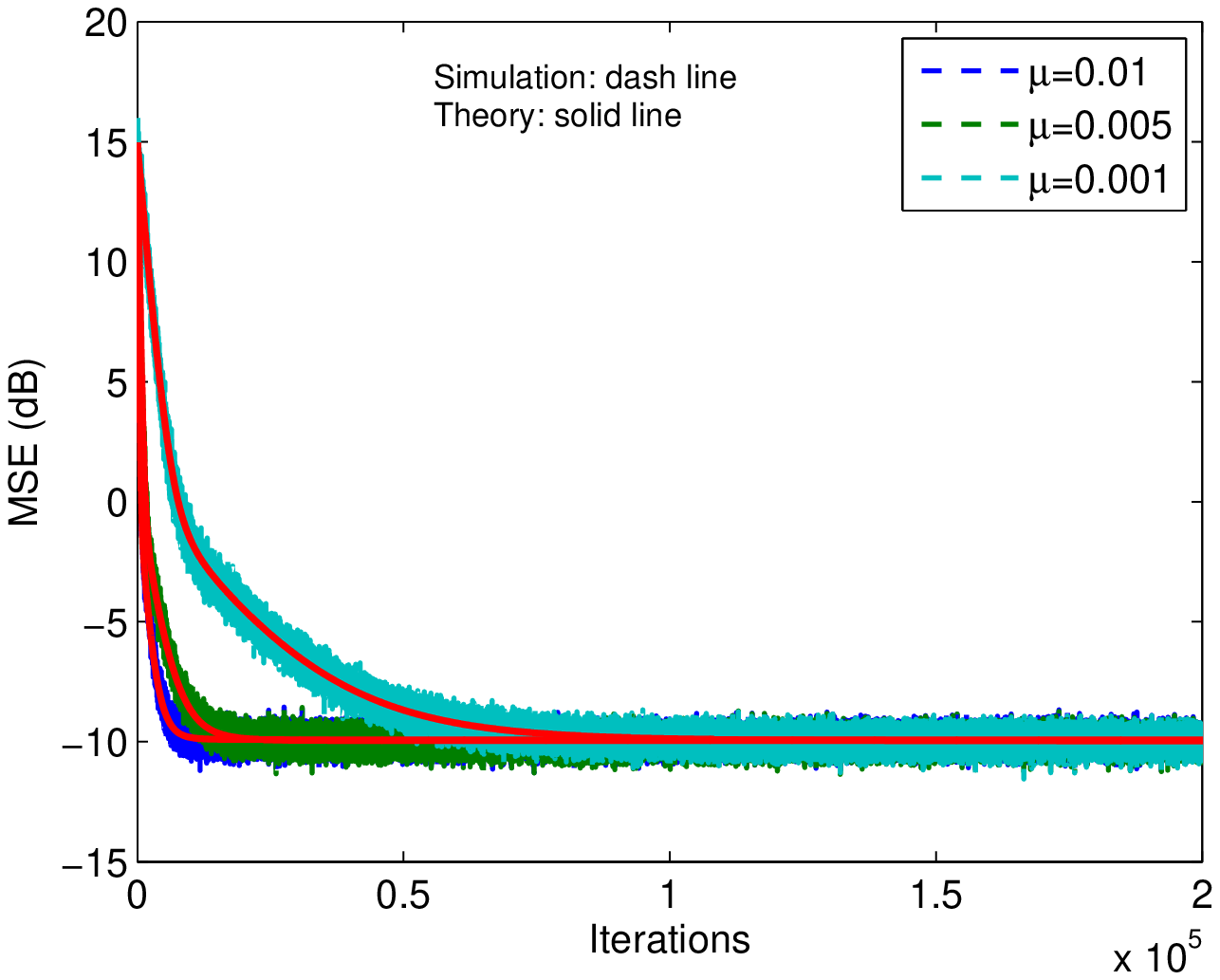}}
\subfigure[]{\includegraphics[height=60mm]{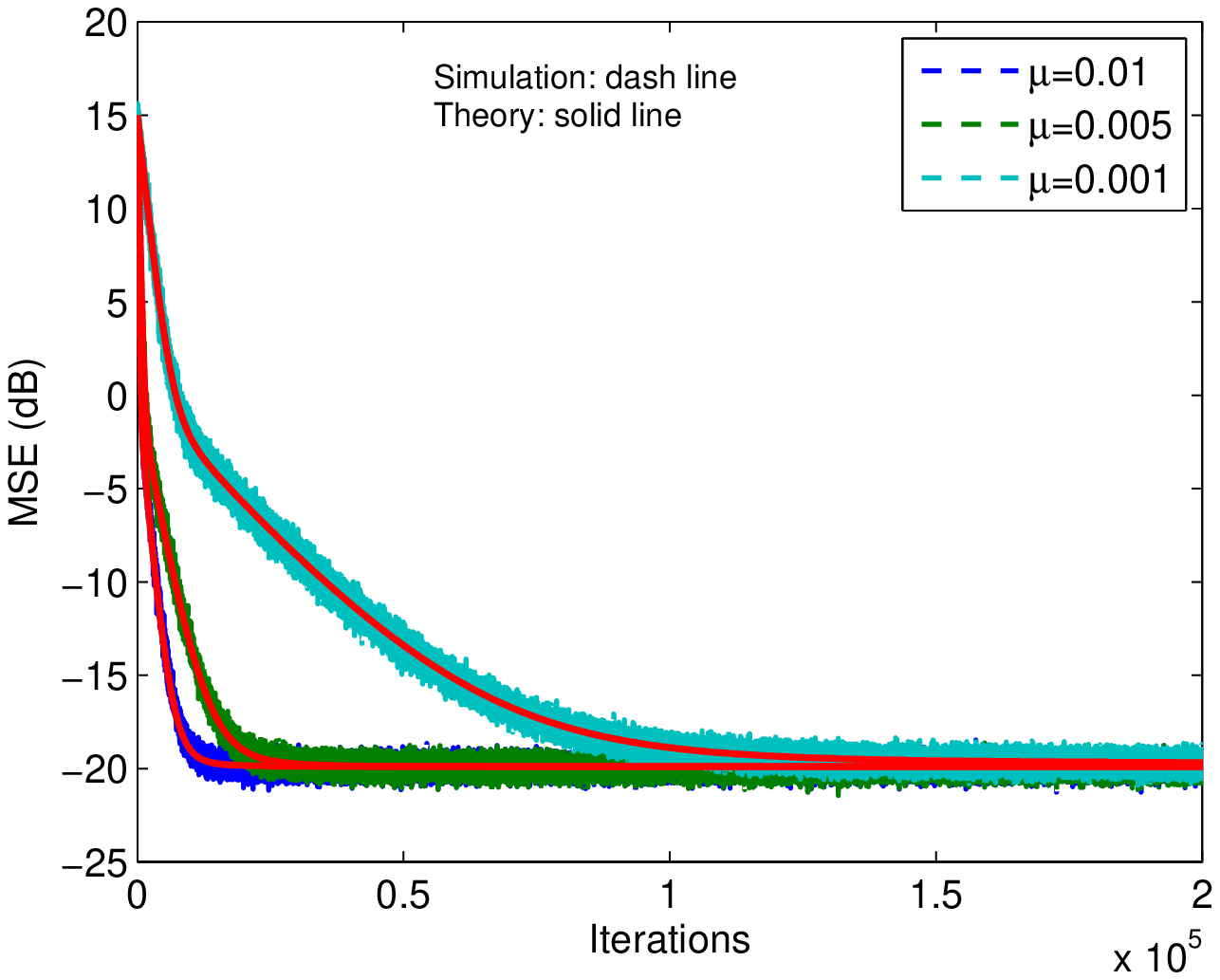}}
\vspace{-1.5ex}
\caption{The MSE learning curves for different step-sizes in the non-stationary environment. (a) the noise variance 0.1, (b) the noise variance 0.01.}
\label{Ther1}
\end{figure*}

\begin{figure*}[!htp]
\centering
\hspace*{-1em}
\subfigure[]{\includegraphics[height=60mm]{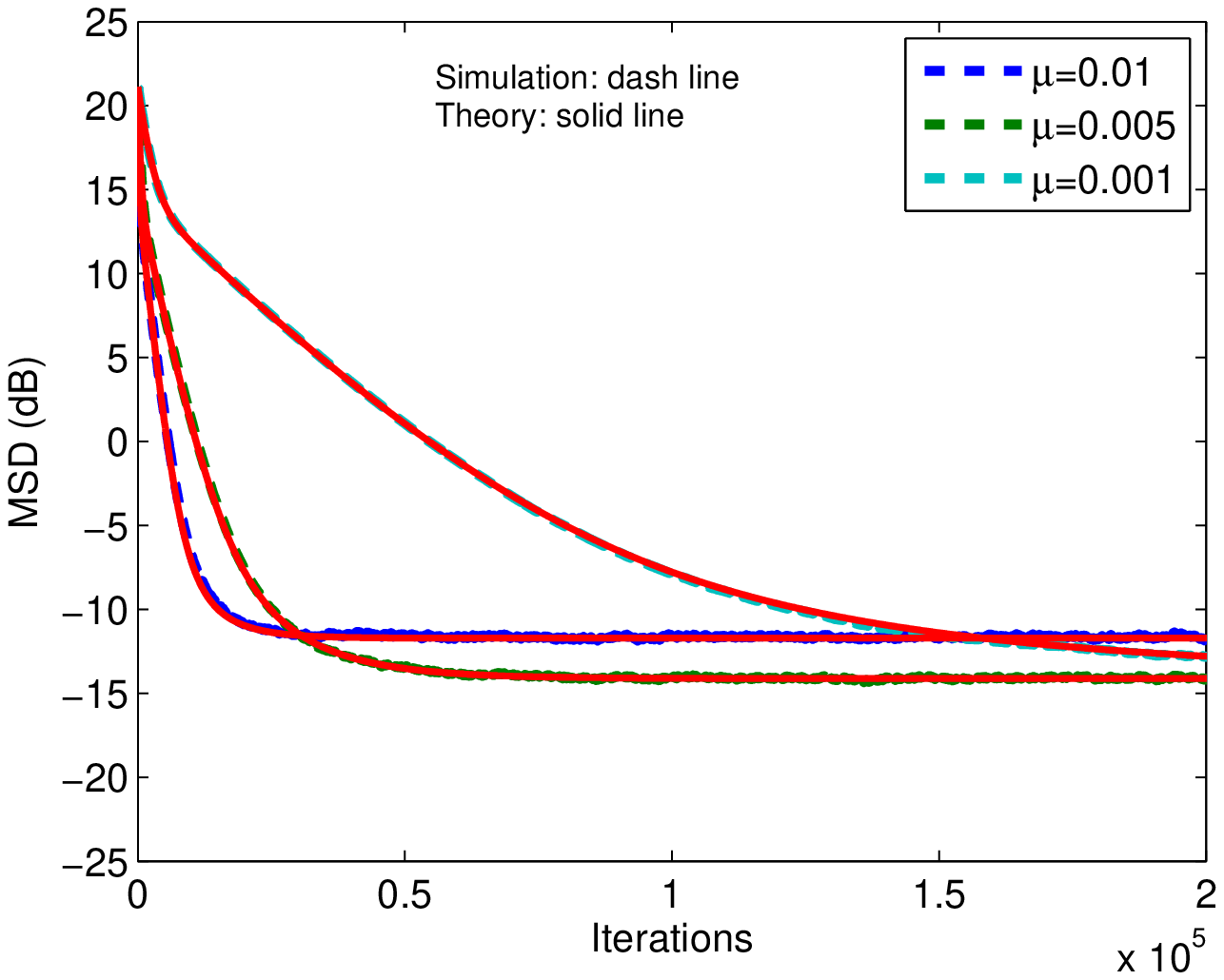}}
\subfigure[]{\includegraphics[height=60mm]{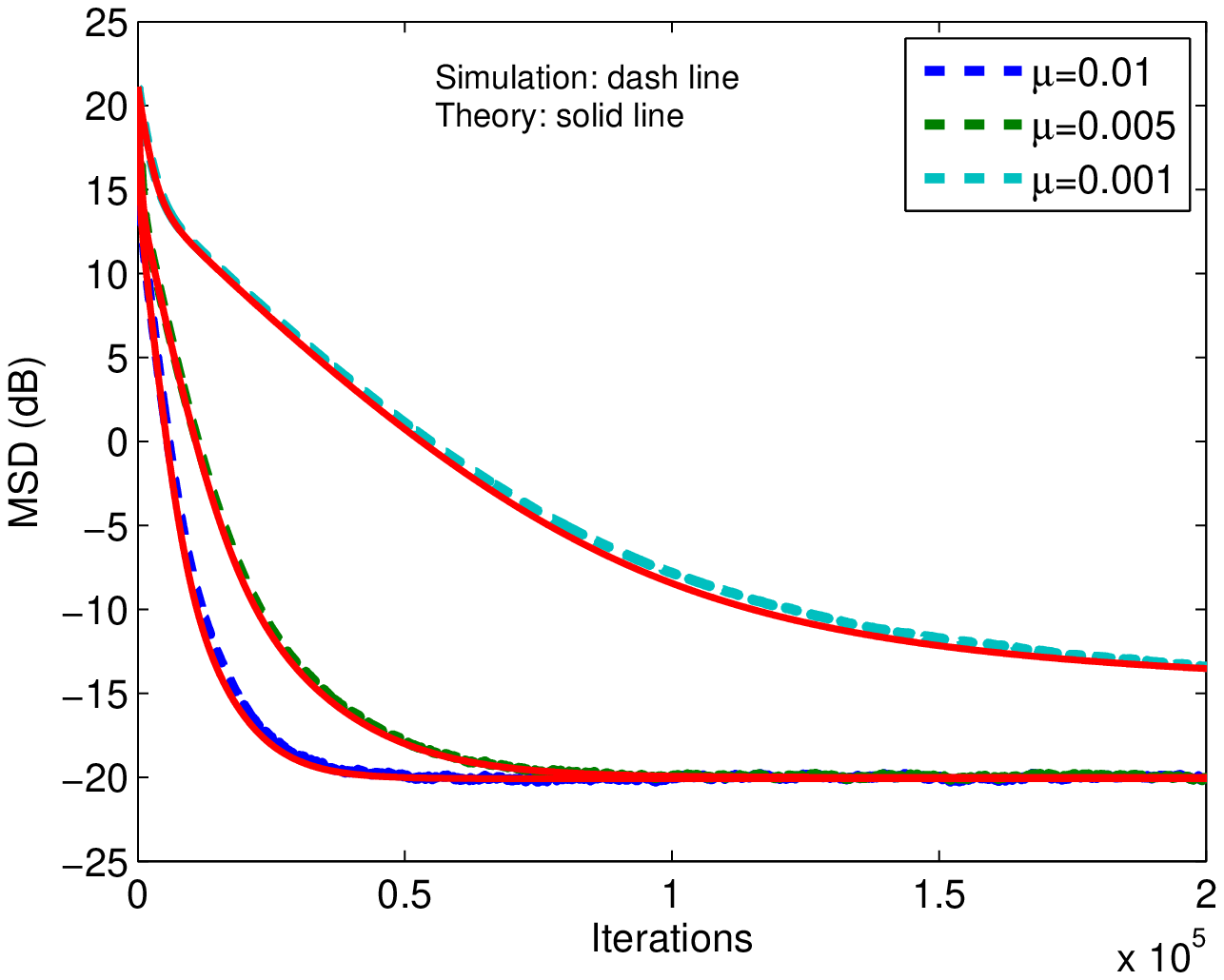}}
\vspace{-1.5ex}
\caption{The MSD learning curves for different step-sizes in the non-stationary environment. (a) the noise variance 0.1, (b) the noise variance 0.01.}
\label{Ther2}
\end{figure*}

To verify the analyses in the section IV, the theoretical transient MSE and MSD curves of the proposed WLRECF for the non-stationary environment ($\sigma^2_q=10^{-8}$) are plotted in Fig.~\ref{Ther1} and Fig.~\ref{Ther2}. According to the model (\ref{12pug234drt2}), the unknown channel is randomly generated and its length is 128. The initial weight vector of the adaptive filter is an all one vector. The input signal $\textbf{x}(n)\in\mathcal{C}^5$ is generated by means of (\ref{esjgert4532}). The vector $\textbf{c}_i$ in (\ref{ekyu5}) is drawn from a white Gaussian distribution with $D=64,\sigma^2=0.2$. Three different step-sizes $\mu = 0.01, 0.005$ and $0.001$ are applied. The variances of the measurement noise $\upsilon_n$ are set to be $0.1,~0.01$ in Fig.~\ref{Ther1} and Fig.~\ref{Ther2}, respectively. The theoretical curves are calculated using (\ref{eq5w4v467862}) and (\ref{eq454687}).
It can be observed that the theoretical analysis can predict the performance of the WLRECF well.
\section{Conclusion\label{sec:Co}}
In this paper, we proposed two random Euler filters, i.e., LRECF and WLRECF, for complex-valued nonlinear filter. On the basis of the \textit{complexification} of real RKHSs and Bochner's theorem, the LRECF filter was firstly derived. Then, further using the widely-linear model and the Wirtinger's derivative, the WLRECF was also obtained. As compared to the well-known complex-valued KLMS, the proposed random Euler filters enjoy low computational complexity because of the fixed network structures. In addition, theoretical expressions to characterize the transient and steady-state behaviors of proposed schemes were presented in the random-walk non-stationary environment. Through a series of simulations, we finally demonstrated the effectiveness of the proposed methods and theoretical results.





\ifCLASSOPTIONcaptionsoff
  \newpage
\fi


\begin{thebibliography}{99}

\bibitem{IEEEhowto1}
C. F. N. Cowan and P. M. Grant, \emph{Adaptive Filters}. Englewood Cliffs, NJ: Prentice-Hall, 1985.

\bibitem{IEEEhowto2}
S. Haykin, \emph{Adaptive Filter Theory}, 4th ed. Upper Saddle River, NJ: Prentice-Hall, 2002.


\bibitem{IEEETS2010}
T. Adali and S. Haykin, \emph{Adaptive Signal Processing: Next Generation Solutions}. Piscataway, NJ: Wiley-IEEE Press, 2010.

\bibitem{IEEEDV2009}
D. P. Mandic and V. S. L. Goh, \emph{Complex Valued Nonlinear Adaptive Filters: Noncircularity, Widely Linear and Neural Models.} New York:
Wiley, 2009

\bibitem{IEEEWJS2010}
W. Liu, J. C. Principe, and S. Haykin, \emph{Kernel Adaptive Filtering}. New York: Wiley, 2010.

\bibitem{IEEEHJ2009}
H. Zhao and J. Zhang, ``Adaptively combined FIR and functional link artificial neural network equalizer for nonlinear communication channel," \emph{IEEE Trans. Neural Netw.}, vol. 20, no. 4, pp. 665-674, Apr. 2009.

\bibitem{IEEEMD2013}
M. Scarpiniti, D. Comminiello, R. Parisi, and A. Uncini, ``Nonlinear spline adaptive filtering," \emph{Signal Process.}, vol. 93, no. 4, pp. 772-783, Apr. 2013.

\bibitem{IEEEAG2014}
A. Carini and G. L. Sicuranza, ``Recursive even mirror Fourier nonlinear filters and simplified structures," \emph{IEEE Trans. Signal Process.}, vol. 62, no. 24, pp. 6534-6544, Dec. 2014.

\bibitem{IEEESW2017}
S. Zhang, and W. X. Zheng, ``Recursive adaptive sparse exponential functional link neural network for nonlinear AEC in impulsive noise environment," \emph{IEEE Trans. Neural Netw. Learn. Syst.}, vol. PP, no. 99, 2017, Doi: 10.1109/TNNLS.2017.2761259.

\bibitem{IEEEAH2006}
A. Hirose, \emph{Complex-valued neural networks}, Springer-Verlag, Berlin, 2006.

\bibitem{IEEEHS1991}
H. Leung, S. Haykin, ``The complex backpropagation algorithm," \emph{IEEE Trans. Signal Process.}, vol. 39, no. 9, pp. 2101-2104, 1991.

\bibitem{IEEEDJ2001}
D. P. Mandic and J. A. Chambers, \emph{Recurrent Neural Networks for Prediction: Learning Algorithms, Architectures and Stability}. New York:
Wiley, 2001.

\bibitem{IEEEHT2008}
H. Li and T. Adali, ``Complex-valued adaptive signal processing using nonlinear functions," \emph{J. Adv. Signal Process. Special Issue on Emerging
Mach. Learn. Tech. Signal Process.}, 2008, 765 615.

\bibitem{IEEEYBMJ2011}
Y. Xia, B. Jelfs, M. M. V. Hulle, J. C. Pr¨ªncipe, and D. P. Mandic,  ``An augmented echo state network for nonlinear adaptive filtering of complex noncircular signals," \emph{IEEE Trans. neural networks}, vol. 22, no. 1, pp. 74-83, Jan. 2011.

\bibitem{IEEEMJYW2012}
M. Li, J. Liu, Y. Jiang, and W. Feng, ``Complex-chebyshev functional link neural network behavioral model for broadband wireless power amplifiers," \emph{IEEE Trans. Microw. Theory Tech.}, vol. 60, no. 6, pp. 1979-1989, Jun. 2012.

\bibitem{IEEEKSI2009}
K. Slavakis, S. Theodoridis, and I. Yamada, ``Adaptive constrained learning in reproducing kernel Hilbert spaces: The robust beamforming case," \emph{IEEE Trans. Signal Process.}, vol. 57, no. 12, pp. 4744-4764, 2009.

\bibitem{IEEEWJCJ2012}
W. Parreira, J. Bermudez, C. Richard, and J. Tourneret, ``Stochastic behavior analysis of the Gaussian kernel-least-mean-square algorithm," \emph{
IEEE Trans. Signal Process.}, vol. 60, no. 5, p. 2208-2222, 2012.

\bibitem{IEEEWJCJ2004}
W. Liu and J. C. Principe, ``Kernel affine projection algorithms," \emph{EURASIP J. Adv. Signal Process.}, vol. 2008, no. 1, pp. 1-13, Mar. 2008.
\bibitem{IEEEYSR2004}
Y. Engel, S. Mannor, and R. Meir, ``The kernel recursive least-squares algorithm," \emph{IEEE Trans. Signal Process.}, vol. 52, no. 8, 2004.

\bibitem{IEEEPKS2012}
P. Bouboulis, K. Slavakis, and S. Theodoridis, ``Adaptive learning in complex reproducing kernel Hilbert spaces employing Wirtinger's subgradients," \emph{IEEE Trans. Neural Netw. Learn. Syst.}, vol. 23, no. 3, pp. 425-438, 2012.

\bibitem{IEEEFAD2012}
F. A. Tobar, A. Kuh, and D. P. Mandic, ``A novel augmented complex valued kernel LMS," in \emph{Proc. IEEE 7th Sensor Array Multichannel
Signal Process. Workshop (SAM)}, Jun. 2012, pp. 473-476.

\bibitem{IEEEPSM2012}
P. Bouboulis, S. Theodoridis, and M. Mavroforakis, ``The augmented complex kernel LMS," \emph{IEEE Trans. Signal Process.}, vol. 60, no. 9,
pp. 4962-4967, Sept. 2012.

\bibitem{IEEETT2015}
T. K. Paul, and T. Ogunfunmi, ``A kernel adaptive algorithm for quaternion-valued inputs," \emph{IEEE Trans. Neural Netw. Learn. Syst.}, vol. 26, no. 10, Oct. 2015.

\bibitem{IEEEWIJP2009}
W. Liu, I. M. Park, and J. C. Principe, ``An information theoretic approach of designing sparse kernel adaptive filters," \emph{IEEE Trans. Signal
Process.}, vol. 20, no. 12, pp. 1950-1961, Dec. 2009.

\bibitem{IEEEBSPJ2012}
B. Chen, S. Zhao, P. Zhu, and J. Principe, ``Quantized kernel least mean square algorithm," \emph{IEEE Trans. Neural Netw. Learn. Syst.,} vol. 23, no. 1, pp. 22-32, Jan. 2012.

\bibitem{IEEEBSSJ2012}
B. Chen, S. Zhao, S. Seth, and J. C. Principe, ``Online efficient learning with quantized KLMS and $L_1$ regularization," in \emph{Proc. Int. Joint Conf.
Neural Netw. (IJCNN)}, 2012, pp. 1-6.

\bibitem{IEEEANP2012}
A. Singh, N. Ahuja, and P. Moulin, ``Online learning with kernels: Overcoming the growing sum problem," \emph{MLSP}, Sept. 2012.

\bibitem{IEEEWJCJ2014}
W. Gao, J. Chen, C. Richard, and J. Huang, ``Online dictionary learning for kernel LMS," \emph{IEEE Trans. Signal
Process.}, vol. 62, no. 11, pp. 2765-2777, 2014.

\bibitem{IEEEAB2007}
A. Rahimi and B. Recht, ``Random features for large scale kernel machines," in \emph{NIPS}, vol. 20, 2007.

\bibitem{IEEEPSS2016}
P. Bouboulis, S. Pougkakiotis, and S. Theodoridis, ``Efficient KLMS and KRLS algorithms: A random fourier feature perspective," in \emph{Proc. IEEE
Statist. Signal Process. Workshop (SSP),} pp. 1-5, 2016.

\bibitem{IEEEPSS2017}
P. Bouboulis, S. Pougkakiotis, and S. Theodoridis, ``Online distributed learning over networks in RKH spaces using random fourier features,"
arXiv preprint arXiv:1703.08131v2, 2017.

\bibitem{IEEEBP1995}
B. Picinbono and P. Chevalier, ``Widely linear estimation with complex data," \emph{IEEE Trans. Signal Process.}, vol. 43, no. 8, pp. 2030-2033, Aug.
1995.

\bibitem{IEEEYCD2010}
Y. Xia, C. C. Took, and D. P. Mandic, ``An augmented affine projection algorithm for the filtering of complex noncircular signals," \emph{Signal
Process.}, vol. 9, no. 6, pp. 1788-1799, 2010.

\bibitem{IEEEACNR2010}
A. Kuh, C. Manloloyo, N. Corpuz, and R. Kowahl, ``Wind prediction using complex augmented adaptive filters," in \emph{Proc. IEEE Int. Conf.
Green Circuits Syst. (ICGCS)}, 2010, pp. 46-50.

\bibitem{IEEEYLCH2015}
Y.-M. Shi, L. Huang, C. Qian, and H. C. So, ``Shrinkage linear and widely linear complex-valued least mean squares algorithms for adaptive
beamforming," \emph{IEEE Trans. Signal Process.}, vol. 63, no. 1, pp. 119-131, Jan. 2015.

\bibitem{IEEEhowto21}
S. Zhang and J. Zhang, ``Transient analysis of zero attracting NLMS algorithms without Gaussian input signal," \emph{Signal Process.}, vol. 97, pp. 100-109, Apr. 2014.

\bibitem{IEEEhowto22}
S. Zhang and J. Zhang, ``New steady-state analysis results of variable step-size LMS algorithm with different noise distributions," \emph{IEEE Signal Process. Lett.}, vol. 21, no. 6, pp. 653-657, Jun. 2014.

\bibitem{IEEEhowto24}
M. Scarpiniti, D. Comminiello, G. Scarano, R. Parisi, and A. Uncini, ``Steady-state performance of spline adaptive filters," \emph{IEEE Trans. Signal Process.}, vol. 64, no. 4, pp. 816-828, 2016.

\bibitem{IEEEhowto14add}
J. Shi and J. Ni, ``Diffusion sign subband adaptive filtering algorithm with enlarged cooperation and its variant," \emph{Circuits, Systems, and Signal Processing}, vol. 36, no. 4, pp 1714-1724, 2016.

\bibitem{IEEEhowto25}
S. Zhang, J. Zhang, and H. C. So, ``Mean square deviation analysis of LMS and NLMS algorithms with white reference inputs," \emph{Signal Process.}, vol. 131, pp. 20-26, Feb. 2017.

\bibitem{IEEEhowto25add4}
A. H. Sayed, \emph{Fundamentals of Adaptive Filtering}, New York: Wiley-Interscience, 2003.

\bibitem{IEEEISA1995}
I. Cha, and S. A. Kassam, ``Channel equalization using adaptive complex radial basis function networks," \emph{IEEE Journal on Selected Areas in Communications}, vol. 13, no. 1, pp. 122-131, 1995.

\bibitem{IEEEMGBPN2005}
 M. B. Li, G. B. Huang, P. Saratchandran, and N. Sundararajan, ``Fully complex extreme learning machines," \emph{Neurocomputing}, vol. 68, pp. 306-314, 2005.



















\comment{

Normalized Subband Adaptive Filtering Algorithm With Reduced Computational Complexity
By: Petraglia, Mariane R.; Haddad, Diego B.; Marques, Elias L.
IEEE TRANSACTIONS ON CIRCUITS AND SYSTEMS II-EXPRESS BRIEFS   Volume: 62   Issue: 12   Pages: 1164-1168   Published: DEC 2015

Adaptive Consensus for Multiple Nonidentical Matching Nonlinear Systems: An Edge-Based Framework
By: Zhao, Yu; Wen, Guanghui; Duan, Zhisheng; et al.
IEEE TRANSACTIONS ON CIRCUITS AND SYSTEMS II-EXPRESS BRIEFS   Volume: 62   Issue: 1   Pages: 85-89   Published: JAN 2015

Convergence Analysis of an Adaptive Algorithm With Output Power Constraints
By: Kozacky, Walter J.; Ogunfunmi, Tokunbo
IEEE TRANSACTIONS ON CIRCUITS AND SYSTEMS II-EXPRESS BRIEFS   Volume: 61   Issue: 5   Pages: 364-367   Published: MAY 2014

Selective Normalized Subband Adaptive Filter With Subband Extension (vol 60, pg 101, 2013)
By: Song, Moon-Kyu; Kim, Seong-Eun; Choi, Young-Seok; et al.
IEEE TRANSACTIONS ON CIRCUITS AND SYSTEMS II-EXPRESS BRIEFS   Volume: 60   Issue: 7   Pages: 456-456   Published: JUL 2013

A Computationally Efficient Delayless Frequency-Domain Adaptive Filter Algorithm
By: Yang, Feiran; Wu, Ming; Yang, Jun
IEEE TRANSACTIONS ON CIRCUITS AND SYSTEMS II-EXPRESS BRIEFS   Volume: 60   Issue: 4   Pages: 222-226   Published: APR 2013

Selective Normalized Subband Adaptive Filter With Subband Extension
By: Song, Moon-Kyu; Kim, Seong-Eun; Choi, Young-Seok; et al.
IEEE TRANSACTIONS ON CIRCUITS AND SYSTEMS II-EXPRESS BRIEFS   Volume: 60   Issue: 2   Pages: 101-105   Published: FEB 2013

Adaptive Beamforming for Vector-Sensor Arrays Based on a Reweighted Zero-Attracting Quaternion-Valued LMS Algorithm
By: Jiang, Mengdi; Liu, Wei; Li, Yi
IEEE TRANSACTIONS ON CIRCUITS AND SYSTEMS II-EXPRESS BRIEFS   Volume: 63   Issue: 3   Pages: 274-278   Published: MAR 2016

A Proportionate Diffusion LMS Algorithm for Sparse Distributed Estimation
By: Yim, Sung-Hyuk; Lee, Han-Sol; Song, Woo-Jin
IEEE TRANSACTIONS ON CIRCUITS AND SYSTEMS II-EXPRESS BRIEFS   Volume: 62   Issue: 10   Pages: 992-996   Published: OCT 2015

An Improved NLMS Algorithm in Sparse Systems Against Noisy Input Signals
By: Yoo, JinWoo; Shin, JaeWook; Park, PooGyeon
IEEE TRANSACTIONS ON CIRCUITS AND SYSTEMS II-EXPRESS BRIEFS   Volume: 62   Issue: 3   Pages: 271-275   Published: MAR 2015

Set-Membership NLMS Algorithm With Robust Error Bound
By: Zhang, Sheng; Zhang, Jiashu
IEEE TRANSACTIONS ON CIRCUITS AND SYSTEMS II-EXPRESS BRIEFS   Volume: 61   Issue: 7   Pages: 536-540   Published: JUL 2014

}



\end{thebibliography}
\end{document}